%% file: yellowfin_arxiv.tex
\documentclass{article}

\usepackage[verbose=true,letterpaper]{geometry}
\usepackage{microtype}
\usepackage{graphicx}
\usepackage{subfigure}
\usepackage{booktabs} 

\usepackage[draft]{hyperref}
\usepackage[numbers]{natbib}

\usepackage{amsmath,amsbsy,amssymb,amsfonts,amsthm}
\usepackage{bm}
\usepackage{graphicx}
\usepackage{xspace}
\usepackage{color}
\usepackage{algorithm, algpseudocode}
\usepackage{wrapfig}

\usepackage{setspace}
\usepackage{enumitem}
\usepackage{capt-of}

\newtheorem{theorem}{Theorem}
\newtheorem{definition}[theorem]{Definition}

\newtheorem{lemma}[theorem]{Lemma}

\newcommand{\E}{\mathbb{E}}
\newcommand{\Var}{\mathrm{Var}}
\newcommand{\mat}[1]{\bm{\mathit{#1}}}
\algdef{SE}[VARIABLES]{States}{EndStates}
   {\algorithmicvariables}
   {\algorithmicend\ \algorithmicvariables}
\algnewcommand{\algorithmicvariables}{\textbf{States}}
\algrenewcommand\Return{\State \algorithmicreturn{} }

\title{\tuner and the Art of Momentum Tuning}

\author{
  Jian Zhang \\
  Department of Computer Science\\
  Stanford University\\
  \texttt{zjian@stanford.edu}
  \and
  Ioannis Mitliagkas \\
  MILA\\
  University of Montr\'eal\\
  \texttt{ioannis@iro.umontreal.ca}
}

\let\textproc\undefined
\newcommand\textproc{\textsc}
\newcommand{\tuner}{\textsc{YellowFin}\xspace}
\newcommand{\asynctuner}{closed-loop \textsc{YellowFin}\xspace}
\newcommand{\Asynctuner}{Closed-loop \textsc{YellowFin}\xspace}

\newcommand{\yell}[1]{#1}
\newcommand{\outline}[1]{}

\newcommand{\jianedits}[1]{#1}

\begin{document}
\maketitle

\begin{abstract}
\noindent Hyperparameter tuning is one of the most time-consuming workloads in deep learning. 
State-of-the-art optimizers, such as AdaGrad, RMSProp and  Adam,
reduce this labor by adaptively tuning an individual learning rate for each variable.
Recently researchers have shown renewed interest in simpler methods like momentum SGD as they may yield better test metrics.
Motivated by this trend, we ask: can simple adaptive methods based on SGD perform as well or better? We revisit the momentum SGD algorithm and show that hand-tuning a single learning rate and momentum makes it competitive with Adam.
We then analyze its robustness to learning rate misspecification and objective curvature variation.
Based on these insights, we design \tuner, an automatic tuner for momentum and learning rate in SGD.
\tuner optionally uses a negative-feedback loop to compensate for the momentum dynamics in asynchronous settings on the fly.
We empirically show that \tuner can converge in fewer iterations than Adam on ResNets and LSTMs for image recognition, language modeling and constituency parsing,
with a speedup of up to $3.28$x in synchronous and up to $2.69$x in asynchronous settings.
\end{abstract}

\section{Introduction}
Accelerated forms of stochastic gradient descent (SGD), pioneered by
\citet{polyak1964some} and \citet{nesterov1983method}, are the de-facto
training algorithms for deep learning.
Their use requires a sane choice for their {\em hyperparameters}: 
typically a {\em learning rate} and {\em momentum parameter} \citep{sutskever2013importance}.
However, tuning hyperparameters is arguably the most time-consuming part of deep learning, with many papers outlining best tuning practices written
\citep{bengio2012practical,orr2003neural,bengio2012deep,bottou2012stochastic}.
Deep learning researchers have proposed a number of methods to deal with hyperparameter optimization, ranging from grid-search and 
smart black-box methods \citep{bergstra2012random,snoek2012practical}
to adaptive optimizers.
Adaptive optimizers aim to eliminate hyperparameter search by tuning on the fly for a single training run:
algorithms like AdaGrad \citep{duchi2011adaptive}, RMSProp \citep{tieleman2012lecture} and Adam \citep{kingma2014adam} use the magnitude of gradient elements to tune learning rates {\em individually for each variable} and  have been largely successful in relieving practitioners of tuning the learning rate. 

\begin{wrapfigure}[15]{R}{0.55\textwidth}
\vspace{-2.25em}
\begin{minipage}{1.0\linewidth}
\begin{figure}[H]
	\includegraphics[width=0.99\linewidth]{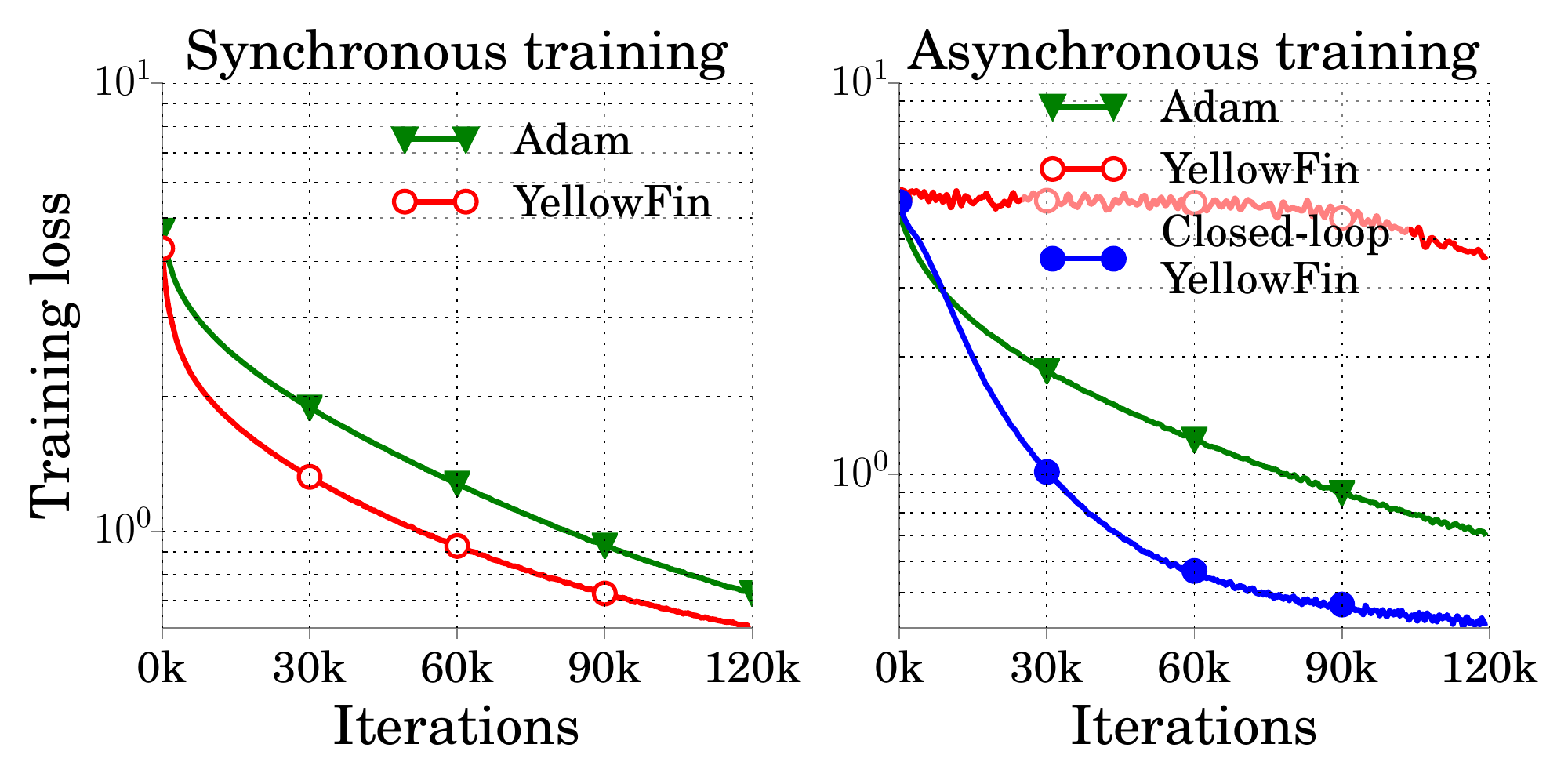}
	\caption{\tuner in comparison to Adam on a ResNet (CIFAR100, cf.\ Section~\ref{sec:experiments}) in synchronous and asynchronous settings.}
	\label{fig:spotlight}
\end{figure}
\end{minipage}
\end{wrapfigure}
Recently some researchers
 have started favoring simple momentum SGD over the previously mentioned adaptive methods~\citep{chen2016thorough,gehring2017convolutional}, often reporting better test scores \citep{wilson2017marginal}.
Motivated by this trend, we ask the question:
\emph{can simpler adaptive methods based on momentum SGD perform as well or better?}
We empirically show, with a hand-tuned learning rate, Polyak's momentum SGD achieves faster convergence than Adam for a large class of models.
We then formulate the optimization update as a dynamical system and study certain robustness properties of the momentum operator.
Inspired by our analysis, we design \tuner, an automatic hyperparameter tuner for momentum SGD.
\tuner simultaneously tunes the learning rate and momentum on the fly, and can handle the complex dynamics of asynchronous execution.
Our contribution and outline are as follows:
\begin{itemize}[leftmargin=2em]
\setlength\itemsep{0.2em}
\item
In Section~\ref{sec:momentum_operator}, we demonstrate examples where momentum offers convergence robust to learning rate misspecification and curvature variation in a class of non-convex objectives.
This robustness is desirable for deep learning.
It stems from a known but obscure fact:
the momentum operator's spectral radius is constant in a large subset of the hyperparameter space.
\item
In Section~\ref{sec:sync_tuner}, we use these robustness insights and a simple quadratic model analysis to motivate the design of \tuner, an automatic tuner for momentum SGD.
\tuner uses on-the-fly measurements from the gradients to tune both a single learning rate and a single momentum.
\item In Section~\ref{sec:stability}, we discuss common stability concerns related to the phenomenon of exploding gradients \citep{pascanu2013difficulty}.
We present a natural extension to our basic tuner, using adaptive gradient clipping, to stabilize training for objectives with exploding gradients.
\item In Section~\ref{sec:async_tuner} we present \asynctuner, suited for asynchronous training.
It uses a novel component for  measuring the total momentum in a running system, including any asynchrony-induced momentum, a phenomenon described in \cite{mitliagkas2016asynchrony}.
This measurement is used in a negative feedback loop to control the value of algorithmic momentum.

\end{itemize}

We provide a thorough empirical evaluation of the performance and stability of our tuner.
In Section~\ref{sec:experiments}, we demonstrate empirically that \yell{on ResNets and LSTMs}
\tuner can converge in fewer iterations compared to:
(i) hand-tuned momentum SGD (up to $1.75$x speedup);
and (ii) hand-tuned Adam ($0.77$x to $3.28$x speedup).
Under asynchrony, the closed-loop control architecture speeds up \tuner, 
making it up to $2.69$x faster than Adam. 
Our experiments include runs on $7$ different models, randomized over at least $3$ different random seeds. 
\tuner is stable and achieves consistent performance: the normalized sample standard deviation of test metrics varies from $0.05\%$ to $0.6\%$.
We released PyTorch and TensorFlow implementations
\footnote{TensorFlow: goo.gl/zC2rjG.  PyTorch: goo.gl/N4sFfs}
that can be used as drop-in replacements for any optimizer.
\tuner has also been implemented in various other packages.
Its large-scale deployment in industry has taught us important lessons about stability; we discuss those challenges and our solution in Section~\ref{sec:stability}.
We conclude with related work and discussion in Section~\ref{sec:related} and~\ref{sec:discussion}.

\vspace{-0.1em}
\section{The momentum operator}
\label{sec:momentum_operator}

\newcommand{\gc}{generalized curvature\xspace}
\newcommand{\Gc}{Generalized curvature\xspace}
\vspace{-0.15em}
In this section, we identify the main technical insight behind the design of \tuner:
 gradient descent with momentum can exhibit linear convergence robust to learning rate misspecification and to curvature variation.
The robustness to learning rate misspecification means tolerance to a less-carefully-tuned learning rate.
On the other hand, the robustness to curvature variation means empirical linear convergence on a class of non-convex objectives with varying curvatures.
After preliminary on momentum, 
we discuss these two properties desirable for deep learning objectives.


\subsection{Preliminaries}
\label{sec:robust_preliminaries}
We aim to minimize some objective $f(x)$.
In machine learning, $x$ is referred to as {\em the model} and the objective is some {\em loss function}.
A low loss implies a well-fit model.
Gradient descent-based procedures use the gradient of the objective function, $\nabla f(x)$, to update the model iteratively.
These procedures can be characterized by the convergence rate with respect to the distance to a minimum.
\begin{definition} [Convergence rate]
	Let $x^*$ be a local minimum of $f(x)$ and $x_t$ denote the model after $t$ steps of an iterative procedure. 
The iterates converge to $x^*$ with linear rate $\beta$,
	if \[ \| x_{t} - x^* \| = O(\beta^t \| x_0 - x^* \|).\]
\end{definition}
Polyak's momentum gradient descent \citep{polyak1964some} is one of these iterative procedures, given by
\begin{align}
	x_{t+1}  &= x_t - \alpha \nabla f(x_t) + \mu (x_t - x_{t-1}),
	\label{eqn:momentum_gd}
\end{align} 
where $\alpha$ denotes a single learning rate and $\mu$ a single momentum for all model variables.   
Momentum's main appeal is its established ability to {accelerate convergence} \citep{polyak1964some}. 
On a $\gamma$-strongly convex $\delta$-smooth function with condition number $\kappa=\delta/\gamma$, the optimal convergence rate of gradient descent without momentum
is $O(\frac{\kappa-1}{\kappa+1})$~\citep{nesterov2013introductory}.
On the other hand, for certain classes of strongly convex and smooth functions, like quadratics,
 the optimal momentum value,
\vspace{-0.5em}
\begin{equation}
	\mu^* = \left(\frac{\sqrt{\kappa}-1}{\sqrt{\kappa}+1}\right)^2,
	\label{eqn:optimal_momentum}
\end{equation}
yields the optimal accelerated linear convergence rate $O(\frac{\sqrt{\kappa}-1}{\sqrt{\kappa}+1})$.
{\em This guarantee does not generalize to arbitrary strongly convex smooth functions} \citep{lessard2016analysis}.
Nonetheless, this linear rate can often be observed in practice even on non-quadratics (cf. Section~\ref{sec:robust_properties}).

{\bf Key insight:}
%
Consider a quadratic objective with condition number $\kappa > 1$.
Even though its curvature is different along the different directions,
Polyak's momentum gradient descent, with $\mu \geq \mu^*$, achieves \emph{the same linear convergence rate $\sqrt{\mu}$ along all directions}. Specifically, let $x_{i, t}$ and $x_i^*$ be the i-th coordinates of $x_t$ and $x^*\!$.
For any $\mu \geq \mu^*$ with an appropriate learning rate, the update in~\eqref{eqn:momentum_gd} can achieve $| x_{i, t} - x_i^* | \leq \sqrt{\mu}^t | x_{i,0} - x_i^* |$ simultaneously along all axes $i$.
This insight has been hidden away in proofs.

In this quadratic case, curvature is different across different axes, but remains {constant on any one-dimensional slice}. 
In the next section (Section~\ref{sec:robust_properties}), we extend this insight to non-quadratic one-dimensional functions.
We then present the \emph{main technical insight behind the design of \tuner: 
similar linear convergence rate $\sqrt{\mu}$ can be achieved in a class of one-dimensional non-convex objectives where curvature varies}; this linear convergence behavior is robust to learning rate misspecification and to the varying curvature. These \emph{robustness properties} are behind a tuning rule for learning rate and momentum in Section~\ref{sec:robust_properties}. We extend this rule to handle SGD noise and generalize it to multidimensional objectives in Section~\ref{sec:sync_tuner}.

\subsection{Robustness properties of the momentum operator}
\label{sec:robust_properties}
In this section, we analyze the dynamics of momentum on a class of one-dimensional, non-convex objectives.
We first introduce the notion of {\em generalized curvature} and use it to describe the momentum operator.
Then we discuss the robustness properties of the momentum operator.

Curvature along different directions is encoded in the different eigenvalues of the Hessian. 
It is the only feature of a quadratic needed to characterize the convergence of gradient descent. Specifically, gradient descent achieves a linear convergence rate $|1 - \alpha h_c|$ on one-dimensional quadratics with constant curvature $h_c$. 
On one-dimensional \emph{non-quadratic objectives with varying curvature}, this neat characterization is lost.
We can recover it by defining a new kind of ``curvature'' with respect to a specific minimum.

\begin{definition}[\Gc]
\label{def:generalized_curvature}
Let $x^*$ be a local minimum of $f(x):\mathbb{R}\rightarrow\mathbb{R}$.
Generalized curvature with respect to $x^*$, denoted by $h(x)$, satisfies the following.
\begin{equation}
	 f'(x) = h(x) (x - x^*). 
	\label{eqn:generalized_curvature}
\end{equation}
\end{definition}
\Gc describes, in some sense, \emph{non-local curvature} with respect to minimum $x^*$.
It coincides with curvature on quadratics.
On non-quadratic objectives, it characterizes the convergence behavior of gradient descent-based algorithms.
Specifically, we recover the fact that starting at point $x_t$, distance from minimum $x^*$ is reduced by $|1-\alpha h(x_t)|$ in one step of gradient descent.
Using a state-space augmentation, we can rewrite the momentum update of~\eqref{eqn:momentum_gd} as
\begin{equation}
\begin{aligned}
{\begin{pmatrix}
x_{t+1} - x^*\\
x_t - x^* \\
\end{pmatrix}}
&= \mat{A}_t
{\begin{pmatrix}
x_t - x^* \\
x_{t-1} - x^*\\
\end{pmatrix}} 
\label{equ:one_dim_22_rec}
\end{aligned}
\end{equation}
where the {\em momentum operator} $\mat{A}_t$ at time $t$ is defined as
\begin{equation}
	\mat{A}_t \triangleq {\begin{bmatrix}
	1-\alpha h(x_t) + \mu & - \mu \\
	1 & 0 \\
	\end{bmatrix}}
\end{equation}

\begin{lemma}[Robustness of the momentum operator]
\label{lem:robustness}
Assume that generalized curvature $h$ and hyperparameters $\alpha,\mu$ satisfy
\begin{align}
{(1-\sqrt{\mu})^2} &\leq \alpha h(x_t) \leq {(1+\sqrt{\mu})^2}.
\label{eqn:robust_region}
\end{align}
Then as proven in Appendix~\ref{sec:proof_robustness}, the spectral radius of the momentum operator at step $t$ depends solely on the  momentum parameter: $	\rho(\mat{A}_t) = \sqrt{\mu}$, for all $t$. 
The inequalities in \eqref{eqn:robust_region} define the {\bf robust region}, the set of learning rate $\alpha$ and momentum $\mu$ achieving this $\sqrt{\mu}$ spectral radius.
\end{lemma}

We know that the spectral radius of an operator, $\mat{A}$, describes its asymptotic behavior when applied multiple times: $\| A^t x \| \approx O(\rho(\mat{A})^t)$.\footnote{
For any $\epsilon > 0$, there exists a matrix norm $\|\cdot\|$ such that $\|\mat{A}\| \leq \rho(A) + \epsilon$~\citep{simon2012spectralradius}.
}
Unfortunately, the same does not always hold for the composition of {\em different } operators, even if they have the same spectral radius, $\rho(\mat{A}_t)=\sqrt{\mu}$.
It is not always true that $\| \mat{A}_t\cdots\mat{A}_1 x\| = O(\sqrt{\mu}^t)$.
However, a homogeneous spectral radius often yields the $\sqrt{\mu}^t$ rate empirically.
In other words, {\em this linear convergence rate is not guaranteed}.
Instead, we demonstrate examples to expose the {\bf robustness properties}: \emph{if the learning rate $\alpha$ and momentum $\mu$ are in the robust region, 
the homogeneity of spectral radii can empirically yield linear convergence with rate $\sqrt{\mu}$; this behavior is robust with respect to learning rate misspecification and to varying curvature}.

\begin{wrapfigure}[17]{R}{0.425\textwidth}
\vspace{-2.5em}
\begin{minipage}{1.0\linewidth}
\begin{figure}[H]
\centering
  \includegraphics[width=0.825\linewidth]{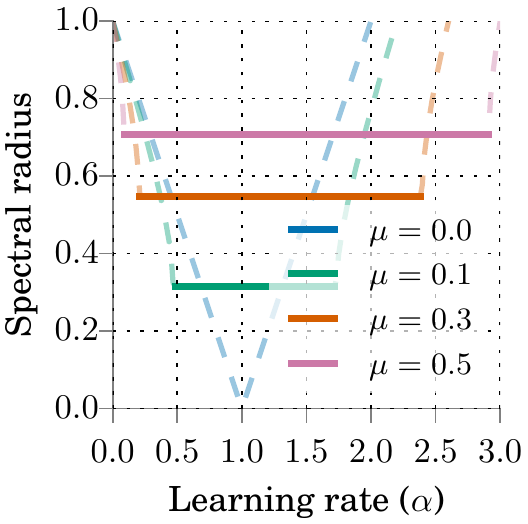}
\caption{
Spectral radius of momentum operator on scalar quadratic
for varying $\alpha$.
}
\label{fig:lr_robustness}
\end{figure}
\end{minipage}
\end{wrapfigure}
\paragraph{Momentum is robust to learning rate misspecification}
\label{sec:lr_robustness}
For a one-dimensional quadratic with curvature $h$,
we have generalized curvature $h(x)=h$ for all $x$. Lemma~\ref{lem:robustness} implies the spectral radius $\rho(\mat{A}_t)\!=\!\sqrt{\mu}$ if
\begin{align}
{(1-\sqrt{\mu})^2/h} &\leq \alpha \leq {(1+\sqrt{\mu})^2/h}.
\label{eqn:lr_robustness}
\end{align}
In Figure~\ref{fig:lr_robustness}, we plot $\rho(\mat{A}_t)$ for different $\alpha$ and $\mu$ when $h\!=\!1$.
The solid line segments correspond to the robust region.
As we increase momentum, a linear rate of convergence, $\sqrt{\mu}$, is robustly achieved by an ever-widening range of learning rates:
higher values of momentum are more robust to learning rate mispecification.
{\bf This property influences the design of our tuner:}
\emph{more generally for a class of one-dimensional non-convex objectives},
as long as the learning rate $\alpha$ and momentum $\mu$ are in the {\em robust region}, i.e.\ satisfy \eqref{eqn:robust_region} at every step, then
{\em momentum operators at all steps $t$ have the same spectral radius}.
In the case of quadratics, this implies a convergence rate of $\sqrt{\mu}$, independent of the learning rate.
Having established that, we can just focus on optimally tuning momentum.

\paragraph{Momentum is robust to varying curvature}
\label{sec:curvature_robustness}

\begin{figure*}[t]
\centering
\vspace{-0.5em}
\begin{tabular}{c c c c}
  \includegraphics[width=0.225\linewidth]{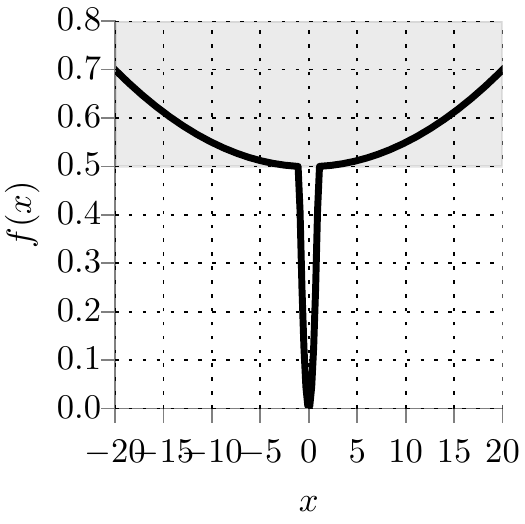} &
  \includegraphics[width=0.235\linewidth]{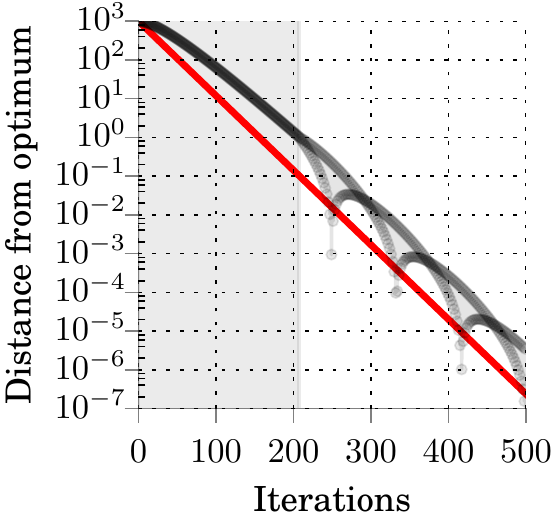} &
  \includegraphics[width=0.235\linewidth]{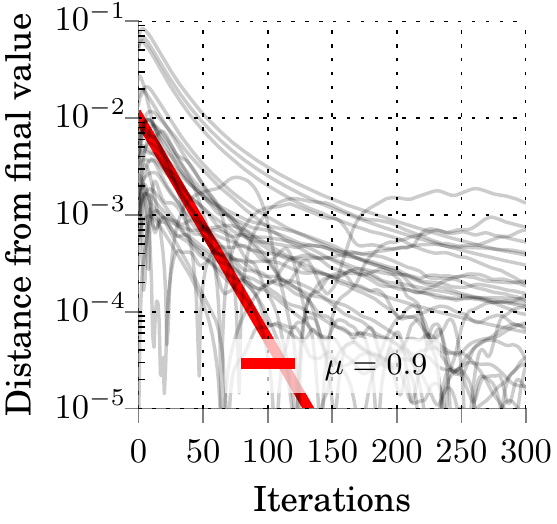} &
  \includegraphics[width=0.185\linewidth]{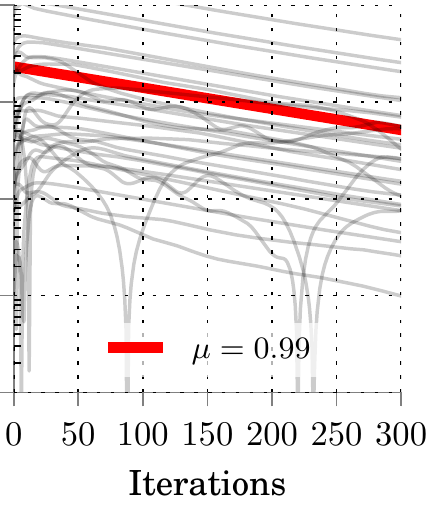} \\
  (a) & (b) & (c) &(d)
\end{tabular}
\vspace{-0.5em}
\caption{(a) Non-convex toy example;
(b) linear convergence rate achieved empirically on the example in (a) tuned according to \eqref{eqn:noiseless_tuning_rule};
(c,d)
LSTM on MNIST: as momentum increases from $0.9$ to $0.99$, the global learning rate and momentum falls in robust regions of more model variables. The convergence behavior (shown in grey) of these variables follow the robust rate $\sqrt{\mu}$ (shown in red).}
\vspace{-0.25em}
\label{fig:curvature_robustness}
\end{figure*}

As discussed in Section~\ref{sec:robust_preliminaries}, the intuition hidden in classic results
is that for certain strongly convex smooth objectives, 
momentum at least as high as
the value in \eqref{eqn:optimal_momentum} can achieve the same rate of linear convergence along all axes with different curvatures. 
We extend this intuition to certain one-dimensional non-convex functions with varying curvatures along their domains; we discuss the generalization to multidimensional cases in Section~\ref{sec:tuner}.
Lemma~\ref{lem:robustness} guarantees constant, time-homogeneous spectral radii for momentum operators $\mat{A}_t$ 
assuming \eqref{eqn:robust_region} is satisfied at every step. 
This assumption motivates a ``long-range'' extension of the condition number.
\begin{definition}[Generalized condition number]
We define the generalized condition number (GCN) with respect to a local minimum $x^*$ of a scalar function, $f(x):\mathbb{R}\rightarrow \mathbb{R}$, to be the dynamic range of its generalized curvature $h(x)$:
\begin{equation}
	\nu = \frac{\sup_{x \in dom(f)} h(x)}{ \inf_{x \in dom(f)} h(x)}
\end{equation}
\end{definition}
The GCN captures variations in generalized curvature along a scalar slice.
From Lemma~\ref{lem:robustness} we get
\begin{equation}
\begin{aligned}
	\mu \geq \mu^* = \left(\frac{\sqrt{\nu}-1}{\sqrt{\nu}+1}\right)^2,
	\quad
	\frac{(1-\sqrt{\mu})^2}{\inf_{x \in dom(f)}h(x)} \leq \alpha \leq \frac{(1+\sqrt{\mu})^2}{\sup_{x \in dom(f)}h(x)}
	\label{eqn:noiseless_tuning_rule}
\end{aligned}
\end{equation}
as the description of the robust region. The momentum and learning rate satisfying~\eqref{eqn:noiseless_tuning_rule} guarantees a homogeneous spectral radius of $\sqrt{\mu}$ for all $\mat{A}_t$.
Specifically, $\mu^*$ is the smallest momentum value that allows for homogeneous spectral radii.
We demonstrate with examples that \emph{homogeneous spectral radii suggest an empirical linear convergence behavior on a class of non-convex objectives}. In Figure~\ref{fig:curvature_robustness}(a), the non-convex objective,
composed of two quadratics with curvatures $1$ and $1000$, has a GCN of $1000$.
Using the tuning rule of \eqref{eqn:noiseless_tuning_rule}, and running the momentum algorithm (Figure~\ref{fig:curvature_robustness}(b)) practically yields the linear convergence predicted by Lemma~\ref{lem:robustness}.
In Figures~\ref{fig:curvature_robustness}(c,d), we demonstrate an LSTM as another example. As we increase the momentum value (the same value for all variables in the model), more model variables follow a $\sqrt{\mu}$ convergence rate.
In these examples, \emph{the linear convergence is robust to the varying curvature of the objectives}. \textbf{This property influences our tuner design:}
{in the next section, we extend the tuning rules of \eqref{eqn:noiseless_tuning_rule} to handle SGD noise; 
we generalize the extended rule to multidimensional cases as the tuning rule in \tuner}.

\input{tuner}

\input{stability}
\input{async_yf}
\input{experiments.tex}
\vspace{-0.5em}
\section{Related work}
\label{sec:related}
\vspace{-0.45em}
Many techniques have been proposed on tuning hyperparameters for optimizers. General hyperparameter tuning approaches, such as random search~\citep{bergstra2012random} and Bayesian approaches~\citep{snoek2012practical, hutter2011sequential}, can directly tune optimizers.  
As another trend, adaptive methods, including AdaGrad~\citep{duchi2011adaptive}, RMSProp~\citep{tieleman2012lecture} and Adam~\citep{kingma2014adam}, uses per-dimension learning rate. 
\citet{schaul2013no} use a noisy quadratic model similar to ours to tune the learning rate in Vanilla SGD.
However they do not use momentum which is essential in training modern neural nets. Existing adaptive momentum approach either consider the deterministic setting~\citep{graepel2002stable,rehman2011effect,hameed2016back,swanston1994simple,ampazis2000levenberg,qiu1992accelerated} or only analyze the stochastic setting with $O(1/t)$ learning rate~\citep{leen1994optimal}. In contrast, we aim at practical momentum adaptivity for stochastically training neural nets.  

\vspace{-0.5em}
\section{Discussion}
\label{sec:discussion}
\vspace{-0.45em}
We presented \tuner, the first optimization method that automatically tunes momentum as well as the learning rate of momentum SGD. 
\tuner outperforms the state-of-the-art adaptive optimizers on a large class of models both in synchronous and asynchronous settings.
It estimates statistics purely from the gradients of a running system,
and then tunes the hyperparameters of momentum SGD based on noisy, local quadratic approximations.
As future work, we believe that more accurate curvature estimation methods,
like the $bbprop$ method~\citep{martens2012estimating} can further improve \tuner.
We also believe that our closed-loop momentum control mechanism in Section~\ref{sec:async_tuner} 
could accelerate other adaptive methods in asynchronous-parallel settings.

\section{Acknowledgements}
We are grateful to Christopher R\'e for his valuable guidance and support. We thank Bryan He, Paroma Varma, Chris De Sa, Tri Dao, Albert Gu, Fred Sala, Alex Ratner and Theodoros Rekatsinas for helpful discussions and feedbacks. We gratefully acknowledge the support of the D3M program under No. FA8750-17-2-0095. Any opinions, findings, and conclusions or recommendations expressed in this material are those of the authors and do not necessarily reflect the views of DARPA or the U.S. government.


\bibliography{arxiv,iclr2018_conference}
\bibliographystyle{unsrtnat}

\appendix 
\input{main_lemma}

\input{opt}
\input{practical_impl}
\input{clip_influence}
\input{async_app}

\input{model_spec}

\input{exp_spec}

%
%
%

\end{document}

%% file: tuner.tex
\vspace{-0.2em}
\section{The \tuner tuner}
\label{sec:sync_tuner}
\vspace{-0.2em}

Here we describe our tuner for momentum SGD that uses the same learning rate for all variables.
We first introduce a noisy quadratic model $f(x)$ as the local approximation of an arbitrary one-dimensional objective. On this approximation, we extend the tuning rule of \eqref{eqn:noiseless_tuning_rule} to SGD. In section~\ref{sec:tuner}, \emph{we generalize the discussion to multidimensional objectives; it yields the \tuner tuning rule}.

\paragraph{Noisy quadratic model}
\label{sec:noisy_quadratics}

\newcommand{\oac}{origin-adjusted curvature }
\newcommand{\bx}{\bar{x}}

We consider a scalar quadratic 
\begin{equation}
	f(x) = \frac{h}{2} x^2 + C 
	= \sum_i \frac{h}{2n}(x-c_i)^2
	\triangleq \frac{1}{n} \sum_i f_i(x)
	\label{eqn:noise_quad_1d}
\end{equation}
with $\sum_i c_i = 0$. $f(x)$ is a quadratic approximation of the original objectives with $h$ and $C$ derived from measurement on the original objective. The function $f(x)$ is defined as the average of $n$ {\em component functions}, $f_i$.
This is a common model for SGD, where we use only a single data point (or a mini-batch) drawn uniformly at random, $S_t \sim \mathrm{Uni}([n])$ to compute a noisy gradient, $\nabla f_{S_t}(x)$, for step $t$.
Here, $C=\frac{1}{2n}\sum_i h c_i^2$ denotes the {\em gradient variance}.
As optimization on quadratics decomposes into scalar problems along the principal eigenvectors of the Hessian, the scalar model in~\eqref{eqn:noise_quad_1d} is sufficient to study local quadratic approximations of multidimensional objectives.
Next we get an {\em exact} expression for the mean square error after running momentum SGD on the scalar quadratic in~\eqref{eqn:noise_quad_1d} for $t$ steps.

\begin{lemma}
\label{lem:main_lemma}
Let $f(x)$ be defined as in \eqref{eqn:noise_quad_1d},
$x_1=x_0$ and $x_t$ follow the momentum update \eqref{eqn:momentum_gd} with stochastic gradients $\nabla f_{S_t}(x_{t-1})$ for $t \geq 2$.
Let $\mat{e}_1=[1, 0]^T$, the expectation of squared distance to the optimum $x^*$ is
	\begin{equation}
	\begin{aligned}
		\E (x_{t+1} - x^{*})^2  = (\mat{e}^{\top}_1 \mat{A}^t [x_1 - x^{*}, x_0-x^{*}]^{\top})^2 
		 + \alpha^2 C \mat{e}^{\top}_1 (\mat{I} - \mat{B}^t)(\mat{I} - \mat{B})^{-1}\mat{e}_1	,
		\label{equ:squared_dist_exact}	
	\end{aligned}
	\end{equation}
where the first and second term correspond to squared bias 
and variance, and their corresponding momentum dynamics are captured by operators
	\begin{equation}
		\mat{A} = \begin{bmatrix}
		1-\alpha h + \mu & - \mu\\
		1 & 0 \\
		\end{bmatrix},
		\quad
		\mat{B} = 
		\begin{bmatrix}
		(1-\alpha h + \mu)^2 &  \mu^2 & -2\mu(1-\alpha h + \mu)\\
		1 & 0 & 0 \\
		1-\alpha h + \mu & 0 & - \mu
		\end{bmatrix}.
		\label{equ:mat_def}
	\end{equation}
\end{lemma}

\yell{
Even though it is possible to numerically work on~\eqref{equ:squared_dist_exact} directly,
we use a scalar, asymptotic surrogate in~\eqref{eqn:asymptotic_surrogate} based on the spectral radii of operators to simplify analysis and expose insights.
This decision is supported by our findings in Section~\ref{sec:momentum_operator}: the spectral radii can capture empirical convergence rate.}
\begin{equation}
\begin{aligned}
	\E ( x_{t+1} - x^{*} )^2 
	\approx  \rho(\mat{A})^{2t} ( x_0 - x_{*} )^2 
		+ (1-\rho(\mat{B})^{t}) \frac{\alpha^2 C}{1-\rho(\mat{B})}
	\label{eqn:asymptotic_surrogate}
\end{aligned}
\end{equation}

One of our design decisions for \tuner 
is to always work in the robust region of Lemma~\ref{lem:robustness}.
We know that this implies a spectral radius $\sqrt{\mu}$ of the momentum operator, $\mat{A}$, for the bias. 
Lemma~\ref{lem:spectral_var_control} shows that under the exact same condition, the variance operator $\mat{B}$ has spectral radius $\mu$.

\begin{lemma}
\label{lem:spectral_var_control}
The spectral radius of the variance operator, $\mat{B}$ is $\mu$, if ${(1-\sqrt{\mu})^2} \leq  \alpha h \leq {(1+\sqrt{\mu})^2}$.
\end{lemma}

As a result, the surrogate objective of \eqref{eqn:asymptotic_surrogate}, takes the following form in the robust region.  
\begin{equation}
	\E ( x_{t+1} - x^{*} )^2 
	\approx \mu^t ( x_0 - x^{*} )^2
		+ (1-\mu^t) \frac{\alpha^2 C}{1-\mu}
	\label{eqn:noisy_square_dist}
\end{equation}
We extend this surrogate to multidimensional cases to extract a noisy tuning rule for \tuner.

\subsection{Tuning rule}
\label{sec:tuner}
\vspace{-0.25em}

In this section, we present \textsc{SingleStep}, the tuning rule of YellowFin (Algorithm~\ref{alg:basic-algo}). Based on the surrogate in~\eqref{eqn:noisy_square_dist}, \textsc{SingleStep} is a multidimensional SGD version of the noiseless tuning rule in~\eqref{eqn:noiseless_tuning_rule}. We first generalize~\eqref{eqn:noiseless_tuning_rule} and~\eqref{eqn:noisy_square_dist} to multidimensional cases, and then discuss \textsc{SingleStep}.

As discussed in Section~\ref{sec:robust_properties}, GCN $\nu$ captures the dynamic range of generalized curvatures in a one-dimensional objective with varying curvature. The consequent robust region described by~\eqref{eqn:noiseless_tuning_rule} implies homogeneous spectral radii. 
On a multidimensional non-convex objective, each one-dimensional slice passing a minimum $x^*$ can have \emph{varying curvature}. As we use \emph{a single $\mu$ and $\alpha$ for the entire model}, if $\nu$ simultaneously captures the dynamic range of generalized curvature over all these slices, $\mu$ and $\alpha$ in~\eqref{eqn:noiseless_tuning_rule} are in the robust region for all these slices. This implies homogeneous spectral radii $\sqrt{\mu}$ according to Lemma~\ref{lem:robustness}, empirically facilitating convergence at a common rate along all the directions. 

Given homogeneous spectral radii $\sqrt{\mu}$ along all directions, the surrogate in~\eqref{eqn:noisy_square_dist} generalizes on the local quadratic approximation of multiple dimensional objectives. On this approximation with minimum $x^*$, the expectation of squared distance to $x^*$, $\E \| x_0 - x^*\|^2$, decomposes into independent scalar components along the eigenvectors of the Hessian. We define gradient variance $C$ as the sum of gradient variance along these eigenvectors. The one-dimensional surrogates in~\eqref{eqn:noisy_square_dist} for the independent components sum to $\mu^t\| x_0 - x^* \|^2 + (1-\mu^t)\alpha^2 C / (1 - \mu)$, the \emph{multidimensional surrogate} corresponding to the one in~\eqref{eqn:noisy_square_dist}. 
\begin{minipage}{0.475\linewidth}
	\begin{equation}
	\begin{aligned}
	&\textsc{(SingleStep)} \\
	 \mu_t, \alpha_t = & \arg \min_{\mu} \mu D^2
		+ \alpha^2 C \\
	s.t.\  \mu \geq & \left(\frac{\sqrt{h_{\max}/h_{\min} }-1}{\sqrt{h_{\max}/h_{\min}}+1}\right)^2 \\
	\alpha =& \frac{(1-\sqrt{\mu})^2}{h_{\min}}
	\end{aligned}
	\label{equ:noisy_min}
	\end{equation}
\end{minipage}
\begin{minipage}{0.025\linewidth}
\end{minipage}
\begin{minipage}{0.475\linewidth}
\begin{algorithm}[H]
	\caption{\jianedits{\tuner}}
	\begin{algorithmic}
	\Function{\tuner}{$\text{gradient } g_t$, $\beta$}
	\State $h_{\max}, h_{\min} \gets \Call{CurvatureRange}{g_t, \beta}$
	\State $C \gets \Call{Variance}{g_t, \beta}$ 
	\State $D \gets \Call{Distance}{g_t, \beta}$ 

	\State $\mu_t, \alpha_t \gets \Call{SingleStep}{C, D, h_{\max}, h_{\min}}$
	\Return $\mu_t, \alpha_t$
	\EndFunction
	\end{algorithmic}
	\label{alg:basic-algo}
\end{algorithm}
\end{minipage}

Let $D$ be an estimate of the current model's distance to a local quadratic approximation's minimum, and $C$ denote an estimate for gradient variance.
\textsc{SingleStep} minimizes the \emph{multidimensional surrogate} after a single step (i.e. $t=1$) while ensuring $\mu$ and $\alpha$ in the robust region for all directions. \emph{A single instance of \textsc{SingleStep} solves a single momentum and learning rate for the entire model at each iteration.}
Specifically, the extremal curvatures $h_{min}$ and $h_{max}$ denote estimates for the largest and smallest generalized curvature respectively. They are meant to capture both generalized curvature variation along all different directions (like the classic condition number)
and also variation that occurs as the {\em landscape evolves}. The constraints keep the global learning rate and momentum in the robust region (defined in Lemma~\ref{lem:robustness}) 
for slices along all directions.
\textsc{SingleStep} can be solved in closed form; we refer to Appendix~\ref{sec:opt} for relevant details on the closed form solution. 
\tuner uses functions \textproc{CurvatureRange}, \textproc{Variance} and \textproc{Distance} to measure quantities $h_{\max}$, $h_{\min}$, $C$ and $D$ respectively. These measurement functions can be designed in different ways.
We present the implementations we used for our experiments,
based completely on gradients,  in Section~\ref{sec:oracles}.

\input{oracles}

%% file: oracles.tex
\subsection{Measurement functions in \tuner}
\label{sec:oracles}
This section describes our implementation of the measurement oracles used by \tuner: \textproc{CurvatureRange}, \textproc{Variance}, and \textproc{Distance}.
We design the measurement functions with the assumption of a negative log-probability objective; this is in line with typical losses in machine learning, e.g. cross-entropy for neural nets and maximum likelihood estimation in general.
Under this assumption, the Fisher information matrix---i.e.\ the expected outer product of noisy gradients---approximates the Hessian of the objective~\citep{johnfisherinfo2016,pascanu2013revisiting}. This allows for measurements purely from minibatch gradients with overhead linear to model dimensionality.
These implementations are not guaranteed to give accurate measurements.
Nonetheless, their use in our experiments in Section~\ref{sec:experiments} shows that they are sufficient for \tuner to outperform the state of the art on a variety of objectives. We also refer to Appendix~\ref{sec:practical_impl} for details on zero-debias~\citep{kingma2014adam}, slow start~\citep{schaul2013no} and smoothing for curvature range estimation.

\begin{table*}[t]
\begin{minipage}{0.37\textwidth}
\vspace{-1em}
\algrenewcommand\alglinenumber[1]{\scriptsize #1:}
	\begin{algorithm}[H]
	\scriptsize
	\setstretch{1.01}
	\caption{\small Curvature range}
	\begin{algorithmic}
		\State \textbf{state: } $h_{\max}$, $h_{\min}$, $h_i, \forall i \in\{1,2,3,...\}$
		\Function{CurvatureRange}{gradient $g_t$, $\beta$}
			\State $h_t \gets \| g_t \|^2$
			\State $h_{\max,t}\gets\!\!\!\max\limits_{t - w \leq i \leq t}\!\!h_i$, $h_{\min,t}\gets\!\!\!\min\limits_{t - w \leq i \leq t}\!\!h_i$
			\State $h_{\max} \gets \beta \cdot h_{\max} + (1 - \beta) \cdot h_{\max,t}$ 
			\State $h_{\min} \gets \beta \cdot h_{\min} + (1 - \beta) \cdot h_{\min,t}$ 
			\Return $h_{\max}$, $h_{\min}$
		\EndFunction
	\end{algorithmic}
	\label{alg:curv_func}
	\end{algorithm}
\end{minipage}
\begin{minipage}{0.315\textwidth}
\vspace{-1em}
\algrenewcommand\alglinenumber[1]{\scriptsize #1:}
	\begin{algorithm}[H]
	\scriptsize
	\setstretch{1.5}
	\caption{\small Gradient variance}
	\begin{algorithmic}
	\State \textbf{state: } $\overline{g^2}\gets0$, $\overline{g}\gets0$
	\Function{Variance}{gradient $g_t$, $\beta$}
		\State $\overline{g^2}\gets\beta \cdot \overline{g^2} + (1 - \beta) \cdot g_t \odot g_t$
		\State $\overline{g}\gets\beta \cdot \overline{g} + (1 - \beta) \cdot g_t$
		\Return $\bm{1}^T\!\!\cdot\left(\overline{g^2} - \overline{g}^2\right)$ 
	\EndFunction
	\end{algorithmic}
	\label{alg:var_func}
	\end{algorithm}
\end{minipage}
\begin{minipage}{0.3\textwidth}
\vspace{-1em}
\algrenewcommand\alglinenumber[1]{\scriptsize #1:}
	\begin{algorithm}[H]
	\scriptsize
	\setstretch{1.25}
	\caption{\small Distance to opt.}
	\begin{algorithmic}
	\State \textbf{state: } $\overline{\|g\|}\gets0$, $\overline{h}\gets0$
		\Function{Distance}{gradient $g_t$, $\beta$}
		\State $\overline{\|g\|}\gets \beta \cdot \overline{\|g\|} + (1 - \beta) \cdot \|g_t\|$
		\State $\overline{h} \gets \beta \cdot \overline{h} + (1 - \beta) \cdot \| g_t \|^2$
		\State $D \gets \beta \cdot D + (1 - \beta) \cdot \overline{\|g\|} /\overline{h}$
		\Return $D$
	\EndFunction
	\end{algorithmic}
	\label{alg:dist_func}
	\end{algorithm}
\end{minipage}
\end{table*}

\paragraph{Curvature range}
Let $g_t$ be a noisy gradient, we estimate the curvatures range in Algorithm~\ref{alg:curv_func}. We notice that the outer product $g_tg_t^T$ has an eigenvalue $h_t=\| g_t \|^2$ with eigenvector $g_t$. Thus under our negative log-likelihood assumption, we use $h_t$ to approximate the curvature of Hessian along gradient direction $g_t$.
Specifically, we maintain $h_{\min}$ and $h_{\max}$ as running averages of extreme curvature $h_{\min, t}$ and $h_{\max, t}$, from a sliding window of width 20.
As gradient directions evolve, we estimate curvatures along different directions. Thus $h_{\min}$ and $h_{\max}$ capture the curvature variations.

\paragraph{Gradient variance}
To estimate the gradient variance in Algorithm~\ref{alg:var_func}, 
we use running averages $\overline{g}$ and $\overline{g^2}$ to keep track of $g_t$ and $g_t \odot g_t$, the first and second order moment of the gradient. 
As $\Var(g_t) = \E{g_t^2} - \E{g_t} \odot \E{g_t}$, we estimate the gradient variance $C$ in \eqref{equ:noisy_min} using $C=\bm{1}^T\!\!\cdot(\overline{g^2} - \overline{g}^2)$. 

\paragraph{Distance to optimum}
In Algorithm~\ref{alg:dist_func}, we estimate the distance to the optimum of the local quadratic approximation.
Inspired by the fact that $\| \nabla f(\mat{x}) \| \leq \| \mat{H} \| \| \mat{x} - \mat{x}^{\star}\|$ for a quadratic $f(x)$ with Hessian $\mat{H}$ and minimizer $\mat{x}^{*}$,  
we first maintain $\overline{h}$ and $\overline{\|g\|}$ as running averages of curvature $h_t$ and gradient norm $\| g_t \|$. Then the distance is approximated using $\overline{\|g\|} / \overline{h}$. 

%% file: stability.tex
\subsection{Stability on non-smooth objectives}
\label{sec:stability}


\begin{wrapfigure}[10]{r}{0.475\linewidth}
\vspace{0.5em}
\begin{minipage}{\linewidth}
\centering
\vspace{-0.75em}
\begin{tabular} {c | c c }
\toprule
	& Loss & BLEU4 \\
\midrule
\midrule
	Default w/o clip. & \multicolumn{2}{c}{diverge} \\ [0.3em]
	Default w/ clip. & 2.86 & 30.75 \\ [0.3em]
	YF & \textbf{2.75} & \textbf{31.59} \\
\bottomrule
\end{tabular}
\captionof{table}{German-English translation validation metrics using convolutional seq-to-seq model.}
\label{tab:conv_seq}
\end{minipage}
\end{wrapfigure}
The process of training neural networks is inherently non-stationary, with the landscape abruptly switching from flat to steep areas. 
In particular, the objective functions of RNNs with hidden units can exhibit occasional but very steep slopes \citep{pascanu2013difficulty,szegedy2013intriguing}.
To deal with this issue, we use \emph{adaptive gradient clipping} heuristics as a very natural addition to our basic tuner. It is discussed with extensive details in Appendix~\ref{sec:adapt_clip}.  
In Figure~\ref{fig:stability} in Appendix~\ref{sec:adapt_clip}, we present an example of an LSTM that exhibits the 'exploding gradient' issue. The proposed adaptive clipping can stabilize the training process using \tuner and prevent large catastrophic loss spikes.

We validate the proposed adaptive clipping on the convolutional sequence to sequence learning model \citep{gehring2017convolutional} for IWSLT 2014 German-English translation. The default optimizer~\citep{gehring2017convolutional} uses learning rate $0.25$ and Nesterov's momentum $0.99$, diverging to loss overflow due to 'exploding gradient'. It requires, as in~\citet{gehring2017convolutional}, strict manually set gradient norm threshold $0.1$ to stabilize. 
In Table~\ref{tab:conv_seq}, we can see YellowFin, with adaptive clipping, outperforms the default optimizer using manually set clipping, with 0.84 higher validation BLEU4 after 120 epochs.

%% file: async_yf.tex
\begin{figure}
\vspace{-1em}
\centering
\includegraphics[width=0.99\linewidth]{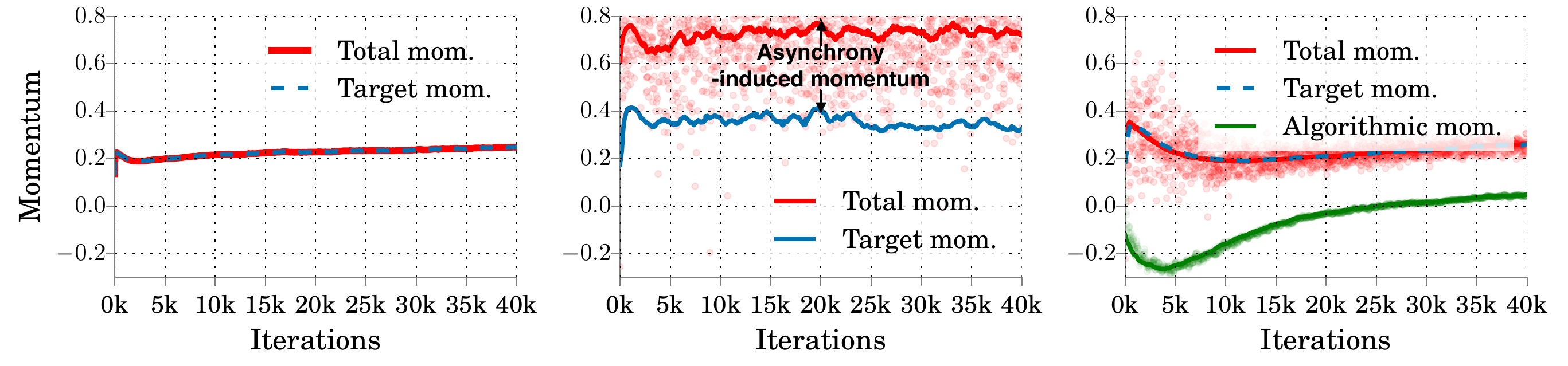}
	\caption{
	When running \tuner, total momentum $\hat{\mu}_t$ equals algorithmic value in synchronous settings (left); $\hat{\mu}_t$ is greater than algorithmic value on 16 asynchronous workers (middle).
	\Asynctuner automatically lowers algorithmic momentum and brings total momentum to match the target value (right).
	Red dots are total momentum estimates, $\hat{\mu}_T$, at each iteration. 
The solid red line is a running average of $\hat{\mu}_T$.
	}
	\label{fig:we-can-measure}
\end{figure}

\section{\Asynctuner}
\label{sec:async_tuner}
Asynchrony is a parallelization technique that avoids synchronization barriers \citep{recht2011hogwild}. 
It yields better hardware efficiency, i.e. faster steps, but can
increase the number of iterations to a given metric, i.e. statistical efficiency, as a tradeoff~\citep{DBLP:journals/pvldb/ZhangR14}.
\citet{mitliagkas2016asynchrony} interpret asynchrony as added momentum dynamics.
We design \asynctuner, a variant of \tuner to automatically control algorithmic momentum, compensate for asynchrony and accelerate convergence.
We use the formula in~\eqref{equ:exp_async_update} to model the dynamics in the system, where the total momentum, $\mu_T$, includes both asynchrony-induced and algorithmic  momentum, $\mu$, in~\eqref{eqn:momentum_gd}.
\begin{equation}
	\mathbb{E}[ x_{t+1} - x_t ] 
	= \mu_T \mathbb{E}[x_t - x_{t-1}] - \alpha \mathbb{E}\nabla f(x_{t})
\label{equ:exp_async_update}
\end{equation}
We first use~\eqref{equ:exp_async_update} to design an robust estimator $\hat{\mu}_T$ for the value of total momentum at every iteration.
Then we use a simple negative feedback control loop to adjust the value of algorithmic momentum so that $\hat{\mu}_T$ matches the \emph{target momentum} decided by \tuner in Algorithm~\ref{alg:basic-algo}. 
In Figure~\ref{fig:we-can-measure}, 
we demonstrate momentum dynamics in an asynchronous training system. 
As directly using the target value as algorithmic momentum, \tuner (middle) presents total momentum $\hat{\mu}_T$ strictly larger than the target momentum, due to asynchrony-induced momentum. \Asynctuner (right) automatically brings down algorithmic momentum, match measured total momentum $\hat{\mu}_T$ to target value and, as we will see, speeds up convergence comparing to \tuner. We refer to Appendix~\ref{sec:async_app} for details on estimator $\hat{\mu}_T$ and \Asynctuner in Algorithm~\ref{alg:async-algo}.

%% file: experiments.tex
\vspace{-0.25em}
\section{Experiments}
\label{sec:experiments}
\vspace{-0.25em}
We empirically validate the importance of momentum tuning and evaluate \tuner in both synchronous (single-node) and asynchronous settings.
In synchronous settings, we first demonstrate that, with hand-tuning, momentum SGD is competitive with Adam, a state-of-the-art adaptive method.
Then, we evaluate \tuner \emph{without any hand tuning} in comparison to hand-tuned Adam and momentum SGD.
In asynchronous settings, we show that \asynctuner accelerates with momentum closed-loop control, significantly outperforming Adam.

We evaluate on convolutional neural networks (CNN) and recurrent neural networks (RNN). For CNN, we train ResNet~\citep{he2016deep} for image recognition on CIFAR10 and CIFAR100~\citep{krizhevsky2014cifar}.
For RNN, we train LSTMs for character-level language modeling with 
the TinyShakespeare (TS) dataset~\citep{karpathy2015visualizing}, word-level language modeling with the Penn TreeBank (PTB) ~\citep{marcus1993building}, and constituency parsing on the Wall Street Journal (WSJ) dataset~\citep{charniakparsing}.
We refer to Table~\ref{tab:model_specification} in Appendix~\ref{sec:model_spec} for model specifications. 
\emph{To eliminate influences of a specific random seed, in our synchronous and asynchronous experiments, the training loss and validation metrics are averaged from 3 runs using different random seeds.}


\subsection{Synchronous experiments}
\label{subsec:sync_exp}
\begin{figure*}[t]
\centering
	\begin{tabular}{c}
		\includegraphics[width=\linewidth]{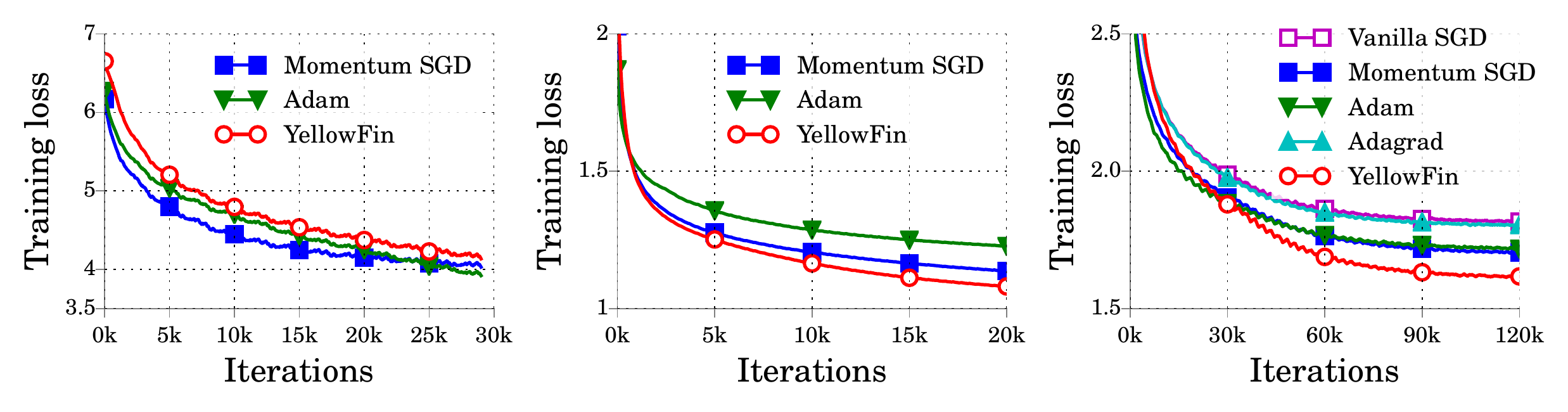} \\[-0.5em]
		\includegraphics[width=\linewidth]{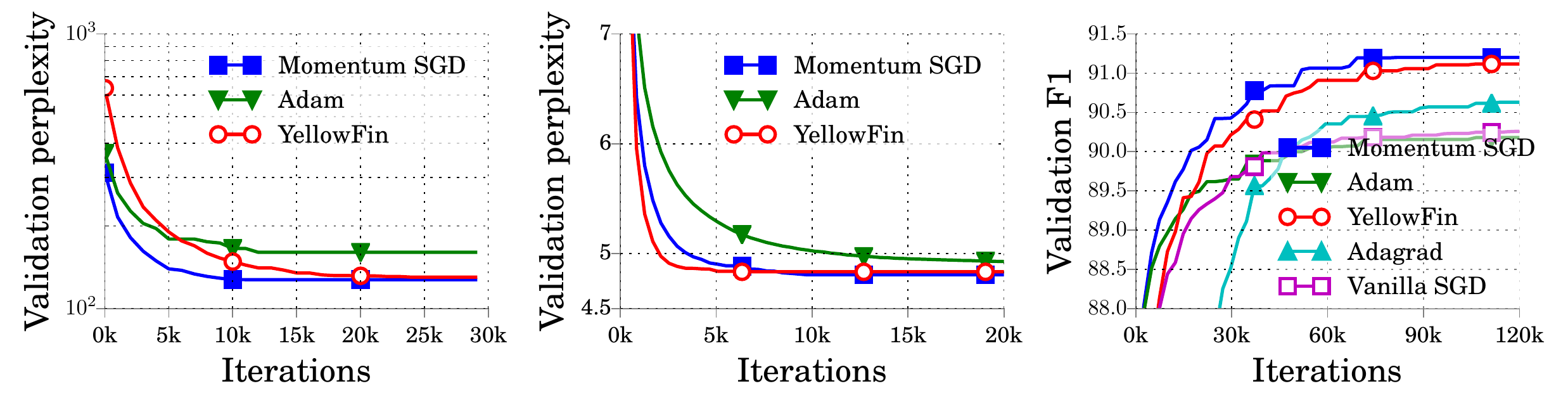} \\[-0.5em]
	\end{tabular}
	\caption{
	Training loss and validation metrics on (left to right) word-level language modeling with PTB, char-level language modeling with TS and constituency parsing on WSJ. The validation metrics are monotonic as we report the best values up to each number of iterations.}
	\label{fig:loss_result_ptb}
\end{figure*}

We tune Adam and  momentum SGD on learning rate grids with prescribed momentum $0.9$ for SGD. We fix the parameters of Algorithm~\ref{alg:basic-algo} in all experiments, i.e.\ \tuner runs {\em without any hand tuning}.
We provide full specifications, including the learning rate (grid) and the number of iterations we train on each model in Appendix~\ref{sec:exp_spec}.
For visualization purposes, we smooth training losses with a uniform window of width $1000$. 
For Adam and momentum SGD on each model, we pick the configuration achieving the lowest averaged smoothed loss.
To compare two algorithms, we record the lowest smoothed loss achieved by both. Then the speedup is reported as the ratio of iterations to achieve this loss.
We use this setup to validate our claims.
\begin{table}[h]
\centering
\small
	\begin{tabular}[t]{@{\hskip 0.5em}c@{\hskip 0.5em}|c@{\hskip 1em}c@{\hskip 1em}c@{\hskip 1em}c@{\hskip 1em}c@{\hskip 0.5em}}
		\toprule
		 & CIFAR10 & CIFAR100 & PTB & TS & WSJ \\
		\midrule
		\midrule
		Adam & 1x & 1x & 1x & 1x & 1x \\
		mom. SGD & 1.71x & 1.87x & 0.88x & 2.49x & 1.33x \\
		YF & 1.93x & 1.38x & 0.77x & 3.28x & 2.33x \\
		\bottomrule
	\end{tabular}
	\caption{
	The speedup of \tuner and tuned momentum SGD over tuned Adam on ResNet and LSTM models.
	}
	\label{tab:iters_to_loss}
\end{table}
\vspace{-0.5em}

\paragraph{Momentum SGD is competitive with adaptive methods}
In Table~\ref{tab:iters_to_loss}, we compare tuned momentum SGD and tuned Adam on ResNets with training losses shown in Figure~\ref{fig:loss_result_cifar} in Appendix~\ref{sec:add_exp}. We can observe that momentum SGD
achieves $1.71$x and $1.87$x speedup to tuned Adam on CIFAR10 and CIFAR100 respectively. In Figure~\ref{fig:loss_result_ptb} and Table~\ref{tab:iters_to_loss}, 
with the exception of PTB LSTM, momentum SGD also produces better training loss, as well as better validation perplexity in language modeling and validation F1 in parsing.
For the parsing task, we also compare with tuned Vanilla SGD and AdaGrad, which are used in the NLP community.
Figure~\ref{fig:loss_result_ptb} (right) shows that \emph{fixed momentum 0.9 can already speedup Vanilla SGD by $2.73$x, achieving observably better validation F1}. 
 We refer to Appendix~\ref{sec:importance_momentum} for further discussion on the importance of momentum adaptivity in \tuner.
\paragraph{\tuner can match hand-tuned momentum SGD and can outperform hand-tuned Adam}
In our experiments, 
\tuner, without any hand-tuning, yields training loss matching hand-tuned momentum SGD for all the ResNet and LSTM models in Figure~\ref{fig:loss_result_ptb} and~\ref{fig:loss_result_cifar}.  
When comparing to tuned Adam in Table~\ref{tab:iters_to_loss}, except being slightly slower on PTB LSTM, \tuner achieves $1.38$x to $3.28$x speedups in training losses on the other four models. \emph{More importantly, \tuner consistently shows better validation metrics than tuned Adam in Figure~\ref{fig:loss_result_ptb}}. It demonstrates that \tuner can match tuned momentum SGD and outperform tuned state-of-the-art adaptive optimizers. 
In Appendix~\ref{sec:boost_exp}, we show \tuner further speeding up with finer-grain manual learning rate tuning.
\subsection{Asynchronous experiments}
\label{sec:async_exp}
In this section, we evaluate \asynctuner with focus on the number of iterations to reach a certain solution. 
To that end, we run $16$ asynchronous workers on a single machine and force them to update the model in a round-robin fashion,
i.e. the gradient is delayed for $15$ iterations.
Figure~\ref{fig:spotlight} (right) 
presents training losses on the CIFAR100 ResNet, using \tuner in Algorithm~\ref{alg:basic-algo}, \asynctuner in Algorithm~\ref{alg:async-algo} and Adam with the learning rate achieving the best smoothed loss in Section~\ref{subsec:sync_exp}.
We can observe closed-loop \tuner achieves $20.1$x speedup to \tuner, 
and consequently a $2.69$x speedup to Adam.
This demonstrates that (1) \asynctuner accelerates by reducing algorithmic momentum to compensate for asynchrony and (2) can converge in less iterations than Adam in asynchronous-parallel training. 

%% file: main_lemma.tex
\newpage
\onecolumn
\section{Proof of Lemma~\ref{lem:robustness}}
\label{sec:proof_robustness}
To prove Lemma~\ref{lem:robustness}, we first prove a more generalized version in Lemma~\ref{lem:robustness_general}. By restricting $f$ to be a one dimensional quadratics function, the generalized curvature $h_t$ itself is the only eigenvalue. We can prove Lemma~\ref{lem:robustness} as a straight-forward corollary. Lemma~\ref{lem:robustness_general} also implies, in the multiple dimensional correspondence of~\eqref{equ:one_dim_22_rec}, the spectral radius $\rho(\mat{A}_t)=\sqrt{\mu}$ if the curvature on all eigenvector directions (eigenvalue) satisfies~\eqref{eqn:robust_region}.

\begin{lemma}
\label{lem:robustness_general}
Let the gradients of a function $f$ be described by
\begin{equation}
	\nabla f(\mat{x}_t) = \mat{H}(\mat{x}_t) (\mat{x}_t - \mat{x}^*),
\end{equation}
with $\mat{H}(\bm{x}_t) \in \mathbb{R}^n \mapsto \mathbb{R}^{n\times n}$.
Then the momentum update can be expressed as a linear operator:
\begin{align}
{\begin{pmatrix}
\mat{y}_{t+1}\\
\mat{y}_t \\
\end{pmatrix}}
=
{\begin{pmatrix}
\mat{I}-\alpha \mat{H}(\mat{x}_t) + \mu \mat{I} & - \mu \mat{I} \\
\mat{I} & \mat{0} \\
\end{pmatrix}}
{\begin{pmatrix}
\mat{y}_t \\
\mat{y}_{t-1} \\
\end{pmatrix}}
=\mat{A}_t
{\begin{pmatrix}
\mat{y}_t \\
\mat{y}_{t-1} \\
\end{pmatrix}},
\end{align}
where $\mat{y}_t\triangleq \mat{x}_t - \mat{x}^*$.
Now, assume that the following condition holds for all eigenvalues $\lambda(\mat{H}(\bm{x}_t))$ of $\mat{H}(\bm{x}_t)$:
\begin{align}
{(1-\sqrt{\mu})^2\over \alpha} &\leq \lambda(\mat{H}(\bm{x}_t)) \leq {(1+\sqrt{\mu})^2\over \alpha}.
\label{equ:control_condition}
\end{align}
then the spectral radius of $\mat{A}_t$ is controlled by momentum with
$	\rho(\mat{A}_t) = \sqrt{\mu}.$

\begin{proof}
Let $\lambda_t$ be an eigenvalue of matrix $\mat{A}_t$, it gives 
$\det\left(\mat{A}_t - \lambda_t \mat{I} \right) = 0$. 
We define the blocks in $\mat{A}_t$ as $\mat{C} = \mat{I} - \alpha \mat{H}_t + \mu \mat{I} - \lambda_t \mat{I}$, $\mat{D} = -\mu \mat{I}$,
$\mat{E} = \mat{I}$ and $\mat{F} = -\lambda_t \mat{I}$ which gives
\[
\det \left( \mat{A}_t - \lambda_t \mat{I}\right) = \det{\mat{F}} \det{\left(\mat{C} - \mat{D} \mat{F}^{-1}
\mat{E} \right)} = 0
\]
assuming generally $\mat{F}$ is invertible. Note we use $\mat{H}_t\triangleq\mat{H}(\mat{x}_t)$ for simplicity in writing. The equation $\det{\left(\mat{C} - \mat{D} \mat{F}^{-1}
\mat{E} \right)} = 0$ implies that
\begin{equation}
\det \left( \lambda_t^2\mat{I} - \lambda_t \mat{M}_t + \mu \mat{I} \right) = 0
\label{equ:control_condition_2}
\end{equation}
with $\mat{M}_t = \left( \mat{I} - \alpha \mat{H}_t + \mu \mat{I} \right)$. In other words, $\lambda_t$ satisfied that $\lambda_t^2 - \lambda_t \lambda(\mat{M}_t) + \mu = 0$ with $\lambda(\mat{M}_t)$ being one eigenvalue of $\mat{M_t}$. I.e.
\begin{equation}
	\lambda_t = \frac{\lambda(\mat{M}_t) \pm \sqrt{\lambda(\mat{M}_t)^2 - 4\mu}}{2}
\end{equation}

On the other hand,~\eqref{equ:control_condition} guarantees that $(1 - \alpha \lambda(\mat{H}_t) + \mu)^2 \leq 4\mu$. We know both $\mat{H}_t$ and $\mat{I} - \alpha \mat{H}_t + \mu \mat{I}$ are symmetric. Thus for all eigenvalues $\lambda(\mat{M}_t)$ of $\mat{M}_t$, we have $\lambda(\mat{M}_t)^2 = (1 - \alpha \lambda(\mat{H}_t) + \mu)^2 \leq 4\mu$ which guarantees $| \lambda_t | = \sqrt{\mu}$ for all $\lambda_t$. As the spectral radius is equal to the magnitude of the largest eigenvalue of $\mat{A}_t$, we have the spectral radius of $\mat{A}_t$ being $\sqrt{\mu}$.

\end{proof}
	
\end{lemma}

\section{Proof of Lemma~\ref{lem:main_lemma}}
We first prove Lemma~\ref{lem:bias_rec} and Lemma~\ref{lem:var_rec} as preparation for the proof of Lemma~\ref{lem:main_lemma}. After the proof for one dimensional case, we discuss the trivial generalization to multiple dimensional case.
\begin{lemma}
\label{lem:bias_rec}
	Let the $h$ be the curvature of a one dimensional quadratic function $f$ and $\overline{x}_t = \E x_t$. We assume, without loss of generality, the optimum point of $f$ is $x^{\star}=0$. Then we have the following recurrence
	\begin{equation} 
		\begin{pmatrix}
			\overline{x}_{t + 1} \\
			\overline{x}_t
		\end{pmatrix} = 
		\begin{pmatrix}
			1-\alpha h + \mu & - \mu\\
			1 & 0 \\
		\end{pmatrix}^{t}
		\begin{pmatrix}
			x_1 \\
			x_0
		\end{pmatrix}
		\label{equ:bias_rec}
	\end{equation} 
	\begin{proof}
		From the recurrence of momentum SGD, 
		we have
		\begin{equation*}
			\begin{aligned}
				\E x_{t + 1} = & \E [ x_{t} - \alpha \nabla f_{S_t} (x_t) + \mu (x_t - x_{t - 1} ) ]\\
							= & \E_{x_{t}}	[ x_{t} - \alpha \E_{S_t} \nabla f_{S_t} (x_t) + \mu (x_t - x_{t - 1} ) ] \\
							= & \E_{x_{t}}	[ x_{t} - \alpha h x_t + \mu (x_t - x_{t - 1} ) ] \\
							= & (1 - \alpha h + \mu)\overline{x}_t - \mu\overline{x}_{t - 1}
			\end{aligned}
		\end{equation*}
		By putting the equation in to matrix form,~\eqref{equ:bias_rec} is a straight-forward result from unrolling the recurrence for $t$ times. Note as we set $x_1 = x_0$ with no uncertainty in momentum SGD, we have $[\overline{x}_0, \overline{x}_1] = [x_0, x_1]$.
	\end{proof}
\end{lemma}

\begin{lemma}
\label{lem:var_rec}
	Let $U_t=\E ( x_t - \overline{x}_t ) ^2$ and $V_t= \E (x_t - \overline{x}_t)(x_{t-1} - \overline{x}_{t-1})$ with $\overline{x}_t$ being the expectation of $x_t$. For quadratic function $f(x)$ with curvature $h \in \mathbb{R}$, We have the following recurrence
		\begin{equation} 
		\begin{pmatrix}
			U_{t+1} \\
			U_t \\
			V_{t + 1}
		\end{pmatrix} = 
		(\mat{I} - \mat{B}^{\top})(\mat{I} - \mat{B})^{-1}
		\begin{pmatrix}
			\alpha^2 C \\
			0 \\
			0
		\end{pmatrix}
	\end{equation}
	where 
	\begin{equation}
		\mat{B} = 
		\begin{pmatrix}
		(1-\alpha h + \mu)^2 &  \mu^2 & -2\mu(1-\alpha h + \mu)\\
		1 & 0 & 0 \\
		1-\alpha h + \mu & 0 & - \mu
		\end{pmatrix}
	\end{equation}
	and $C = \E ( \nabla f_{S_t}(x_t) - \nabla f(x_t) )^2$ is the variance of gradient on minibatch $S_t$.
	
	\begin{proof}
		We prove by first deriving the recurrence for $U_t$ and $V_t$ respectively and combining them in to a matrix form. For $U_t$, we have
		\begin{equation}
		\begin{aligned}
			U_{t + 1} = & \E ( x_{t+1} - \overline{x}_{t + 1} )^2\\
			 = & \E ( x_{t} - \alpha \nabla f_{S_t}(x_t) + \mu (x_{t} - x_{t - 1} ) - (1 - \alpha h + \mu) \overline{x}_t + \mu \overline{x}_{t - 1} )^2 \\
			 = & \E ( x_{t} - \alpha \nabla f(x_t) + \mu (x_{t} - x_{t - 1} ) - (1 - \alpha h + \mu) \overline{x}_t + \mu \overline{x}_{t - 1}  + \alpha (\nabla f(x_t) - \nabla f_{S_t}(x_t)) )^2 \\
			 = & \E ( (1 - \alpha h + \mu) (x_t - \overline{x}_t)  - \mu(x_{t - 1} - \overline{x}_{t - 1} ) )^2 + \alpha^2 \E ( \nabla f(x_t) - \nabla f_{S_t}(x_t) )^2 \\
			 = & (1 - \alpha h + \mu)^2 \E ( x_t - \overline{x}_t )^2 -2 \mu (1 - \alpha h + \mu) \E (x_t - \overline{x}_t)(x_{t - 1} - \overline{x}_{t - 1} ) \\
			 & + \mu^2\E ( x_{t-1} - \overline{x}_{t-1} )^2+ \alpha^2 C
		\end{aligned}
		\label{equ:U_term}
		\end{equation}
		where the cross terms cancels due to the fact $\E_{S_t} [\nabla f(x_t) - \nabla f_{S_t}(x_t)]=0$ in the third equality. 
		
		For $V_t$, we can similarly derive
		\begin{equation}
		\begin{aligned}			
			V_t = & \E (x_t - \overline{x}_t) (x_{t-1} - \overline{x}_{t-1} ) \\
			= & \E ( (1 - \alpha h + \mu) (x_{t-1} - \overline{x}_{t-1} ) - \mu (x_{t-2} - \overline{x}_{t-2}) + \alpha (\nabla f(x_t) - \nabla f_{S_t}(x_t) ) ) (x_{t-1} - \overline{x}_{t-1} ) \\
			= & (1 - \alpha h + \mu)\E ( x_{t-1} - \overline{x}_{t-1} )^2 - \mu \E (x_{t-1} - \overline{x}_{t-1})(x_{t-2} - \overline{x}_{t-2})
		\end{aligned}
		\label{equ:V_term}
		\end{equation}
		Again, the term involving $\nabla f(x_t) - \nabla f_{S_t}(x_t)$ cancels in the third equality as a results of $\E_{S_t} [\nabla f(x_t) - \nabla f_{S_t}(x_t)]=0$.~\eqref{equ:U_term} and~\eqref{equ:V_term} can be jointly expressed in the following matrix form
		\begin{equation}
		\begin{aligned}
			\begin{pmatrix}
			U_{t+1} \\
			U_t \\
			V_{t + 1}
		\end{pmatrix}= \mat{B} 
		\begin{pmatrix}
			U_t \\
			U_{t-1} \\
			V_t
		\end{pmatrix} + 
		\begin{pmatrix}
			\alpha^2 C \\
			0 \\
			0
		\end{pmatrix}
		=\sum\limits_{i = 0}^{t-1} \mat{B}^{i} \begin{pmatrix}
			\alpha^2 C \\
			0 \\
			0
		\end{pmatrix} + \mat{B}^t \begin{pmatrix}
			U_1 \\
			U_0 \\
			V_1
		\end{pmatrix}
		= (\mat{I} - \mat{B}^t)(\mat{I} - \mat{B})^{-1}
		\begin{pmatrix}
			\alpha^2 C \\
			0 \\
			0
		\end{pmatrix}.
		\end{aligned}
		\end{equation}
		Note the second term in the second equality is zero because $x_0$ and $x_1$ are deterministic. Thus $U_1\!=\!U_0\!=\!V_1\!=\!0$.
	\end{proof}
\end{lemma}

According to Lemma~\ref{lem:bias_rec} and~\ref{lem:var_rec}, we have $\E ( \overline{x}_t - x^{*} )^2 = (\mat{e}^{\top}_1 \mat{A}^t [x_1, x_0]^{\top})^2$ and $\E ( x_t - \overline{x}_t )^2=\alpha^2 C \mat{e}^{\top}_1 (\mat{I} - \mat{B}^t)(\mat{I} - \mat{B})^{-1}\mat{e}_1$ where $\mat{e}_1 \in \mathbb{R}^n$ has all zero entries but the first dimension. Combining these two terms, we prove Lemma~\ref{lem:main_lemma}. Though the proof here is for one dimensional quadratics, it trivially generalizes to multiple dimensional quadratics. Specifically, we can decompose the quadratics along the eigenvector directions, and then apply Lemma~\ref{lem:main_lemma} to each eigenvector direction using the corresponding curvature $h$ (eigenvalue). By summing quantities in~\eqref{equ:squared_dist_exact} for all eigenvector directions, we can achieve the multiple dimensional correspondence of~\eqref{equ:squared_dist_exact}.

\section{Proof of Lemma~\ref{lem:spectral_var_control}}
Again we first present a proof of a multiple dimensional generalized version of Lemma~\ref{lem:spectral_var_control}. The proof of Lemma~\ref{lem:spectral_var_control} is a one dimensional special case of Lemma~\ref{lem:spectral_var_control_multi}. Lemma~\ref{lem:spectral_var_control_multi} also implies that for multiple dimension quadratics, the corresponding spectral radius $\rho(\mat{B})=\mu$ if ${(1-\sqrt{\mu})^2\over \alpha} \leq h \leq {(1+\sqrt{\mu})^2\over \alpha}$ on all the eigenvector directions with $h$ being the eigenvalue (curvature).
\begin{lemma}
\label{lem:spectral_var_control_multi}
Let $\mat{H}\in\mathbb{R}^{n\times n}$ be a symmetric matrix and $\rho(\mat{B})$ be the spectral radius of matrix 
\begin{equation}
\mat{B} = {\begin{pmatrix}
(\mat{I}-\alpha \mat{H} + \mu \mat{I})^{\top}(\mat{I}-\alpha \mat{H} + \mu \mat{I}) &  \mu^2 \mat{I} & -2\mu(\mat{I}-\alpha \mat{H} + \mu \mat{I})\\
\mat{I} & \mat{0} & \mat{0} \\
\mat{I}-\alpha \mat{H} + \mu \mat{I} & \mat{0} & - \mu \mat{I} 
\end{pmatrix}}
\end{equation}
We have $\rho(\mat{B})=\mu$ if all eigenvalues $\lambda(\mat{H})$ of $\mat{H}$ satisfies
\begin{equation}
{(1-\sqrt{\mu})^2\over \alpha} \leq \lambda(\mat{H}) \leq {(1+\sqrt{\mu})^2\over \alpha}.
\label{equ:control_condition_var}
\end{equation}

\begin{proof}
	Let $\lambda$ be an eigenvalue of matrix $\mat{B}$, it gives 
$\det\left(\mat{B} - \lambda \mat{I} \right) = 0$ which can be alternatively expressed as
\begin{equation}	
\det \left( \mat{B} - \lambda \mat{I}\right) = \det{\mat{F}} \det{\left(\mat{C} - \mat{D} \mat{F}^{-1}
\mat{E} \right)} = 0
\label{equ:control_condition_var_1}
\end{equation}
assuming $\mat{F}$ is invertible, i.e. $\lambda + \mu \neq 0$, where the blocks in $\mat{B}$ 
\begin{equation*}
		\mat{C} = \left( { \begin{array}{c c}
 			\mat{M}^{\top}\mat{M} - \lambda \mat{I} &  \mu^2 \mat{I} \\
 			\mat{I} & - \lambda \mat{I}
 		\end{array} } \right), 
 		\mat{D} = \left( { \begin{array}{c}
 			-2\mu \mat{M} \\
 			\mat{0}
 		\end{array}}\right),
 		\mat{E} = \left( {\begin{array}{c}
 			\mat{M} \\
 			\mat{0}
 		\end{array}} \right)^{\top},
 		\mat{F} = -\mu \mat{I} - \lambda \mat{I}
	\end{equation*}
	with $\mat{M}=\mat{I}-\alpha \mat{H} + \mu \mat{I}$.~\eqref{equ:control_condition_var_1} can be transformed using straight-forward algebra as
	\begin{equation}
		\det \left( \begin{array}{c c}
 			(\lambda - \mu) \mat{M}^{\top}\mat{M} - (\lambda + \mu) \lambda \mat{I} & (\lambda + \mu)\mu^2 \mat{I} \\
 			(\lambda + \mu) \mat{I} & -(\lambda + \mu)\lambda \mat{I}
 		\end{array} \right) = 0
		\label{equ:control_condition_var_2}	
	\end{equation}
	Using similar simplification technique as in~\eqref{equ:control_condition_var_1}, we can further simplify into
	\begin{equation}
		(\lambda - \mu)\det \left( (\lambda + \mu)^2 \mat{I} - \lambda \mat{M}^{\top}\mat{M} \right) = 0
	\end{equation}
	if $\lambda \neq \mu$, as $(\lambda + \mu)^2 \mat{I} - \lambda \mat{M}^{\top}\mat{M}$ is diagonalizable, we have $(\lambda + \mu)^2 - \lambda \lambda(\mat{M})^2 = 0$ with $\lambda(\mat{M})$ being an eigenvalue of symmetric $\mat{M}$. The analytic solution to the equation can be explicitly expressed as
	\begin{equation}
		\lambda = \frac{\lambda(\mat{M})^2 - 2\mu \pm \sqrt{(\lambda(\mat{M})^2 - 2\mu)^2 - 4\mu^2}}{2}.
		\label{equ:control_condition_var_3}	
	\end{equation}
	
	When the condition in~\eqref{equ:control_condition_var} holds, we have $\lambda(M)^2=(1 - \alpha \lambda(\mat{H}) + \mu)^2 \leq 4\mu$. One can verify that 
	
	\begin{equation}
		\begin{aligned}
			(\lambda(\mat{M})^2 - 2\mu)^2 - 4\mu^2 & = && (\lambda(\mat{M})^2 - 4\mu)\lambda(\mat{M})^2 \\
			& = &&\left( (1 - \alpha \rho(\mat{H} ) + \mu)^2 - 4\mu\right)\lambda(\mat{M})^2 \\
			& \leq && 0
		\end{aligned}
	\end{equation}
	Thus the roots in~\eqref{equ:control_condition_var_3} are conjugate with $| \lambda | = \mu$. In conclusion, the condition in~\eqref{equ:control_condition_var} can guarantee all the eigenvalues of $\mat{B}$ has magnitude $\mu$. Thus the spectral radius of $\mat{B}$ is controlled by $\mu$.
\end{proof}

\end{lemma}

%% file: opt.tex
\section{Analytical solution to~\eqref{equ:noisy_min}}
\label{sec:opt}
The problem in~\eqref{equ:noisy_min} does not need iterative solver but has an analytical solution. Substituting only the second constraint, the objective becomes $p(x)=x^2D^2 + (1-x)^4/h_{\min}^2C$ with $x=\sqrt{\mu} \in [0, 1)$. By setting the gradient of $p(x)$ to 0, we can get a cubic equation whose root $x=\sqrt{\mu_p}$ can be computed in closed form using Vieta's substitution. As $p(x)$ is uni-modal in $[0, 1)$, the optimizer for \eqref{equ:noisy_min} is exactly the maximum of $\mu_p$ and $(\sqrt{h_{\max}/h_{\min} }-1 )^2 / (\sqrt{h_{\max}/h_{\min}}+1)^2$, the right hand-side of the first constraint in~\eqref{equ:noisy_min}.

%% file: practical_impl.tex
\section{Practical implementation}
\label{sec:practical_impl}
In Section~\ref{sec:oracles}, we discuss estimators for learning rate and momentum tuning in \tuner. In our experiment practice, we have identified a few practical implementation details which are important for improving estimators. Zero-debias is proposed by~\citet{kingma2014adam}, which accelerates the process where exponential average adapts to the level of original quantity in the beginning. We applied zero-debias to all the exponential average quantities involved in our estimators. In some LSTM models, we observe that our estimated curvature may decrease quickly along the optimization process. In order to better estimate extremal curvature $h_{\max}$ and $h_{\min}$ with fast decreasing trend, we apply zero-debias exponential average on the logarithmic of $h_{\max, t}$ and $h_{\min, t}$, instead of directly on $h_{\max, t}$ and $h_{\min, t}$. Except from the above two techniques, we also implemented the slow start heuristic proposed by~\citep{schaul2013no}. More specifically, we use $\alpha = \min\{\alpha_t, t \cdot \alpha_t / (10 \cdot w) \}$ as our learning rate with $w$ as the size of our sliding window in $h_{\max}$ and $h_{\min}$ estimation. It discount the learning rate in the first $10 \cdot w$ steps and helps to keep the learning rate small in the beginning when the exponential averaged quantities are not accurate enough.

%% file: clip_influence.tex
\section{Adaptive gradient clipping in \tuner}
\label{sec:adapt_clip}
\begin{figure}
\centering
  \includegraphics[width=0.8\linewidth]{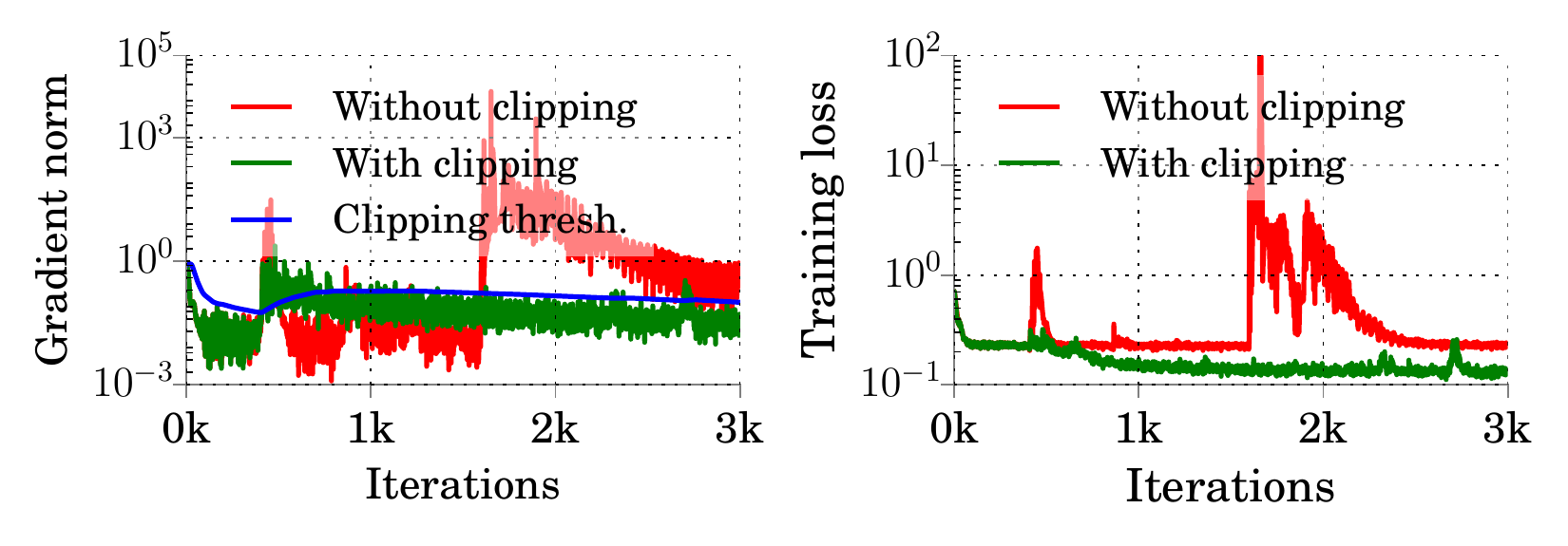} 
\caption{A variation of the LSTM architecture in \citep{zhu2016trained} exhibits exploding gradients.
The proposed adaptive gradient clipping threshold (blue) stabilizes the training loss.}
\label{fig:stability}
\end{figure}

Gradient clipping has been established in literature as a standard---almost necessary---tool for training such objectives \citep{pascanu2013difficulty,Goodfellow-et-al-2016,gehring2017convolutional}. 
However, the classic tradeoff between adaptivity and stability applies: 
setting a clipping threshold that is too low can hurt performance;
setting it to be high, can compromise stability.
\tuner, keeps running estimates of extremal gradient magnitude squares, $h_{max}$ and $h_{min}$ in order to estimate a generalized condition number.
We posit that $\sqrt{h_{max}}$ is an ideal gradient norm threshold for adaptive clipping.
In order to ensure robustness to extreme gradient spikes, like the ones in Figure~\ref{fig:stability}, we also limit the growth rate of the envelope $h_{max}$ in Algorithm~\ref{alg:curv_func} as follows:
\begin{equation}
 h_{max} 
 \leftarrow
 \beta \cdot h_{max}
 	+ (1-\beta) \cdot \textrm{min}\left\{
 		h_{max,t}, 100 \cdot h_{max}
 	\right\}
\end{equation}
Our heuristics follows along the lines of classic recipes like~\cite{pascanu2013difficulty}. However, instead of using the average gradient norm to clip, it uses a running estimate of the maximum norm $h_{\max}$.

\begin{figure}
\centering
\begin{tabular}{c c}
	\includegraphics[width=0.35\linewidth]{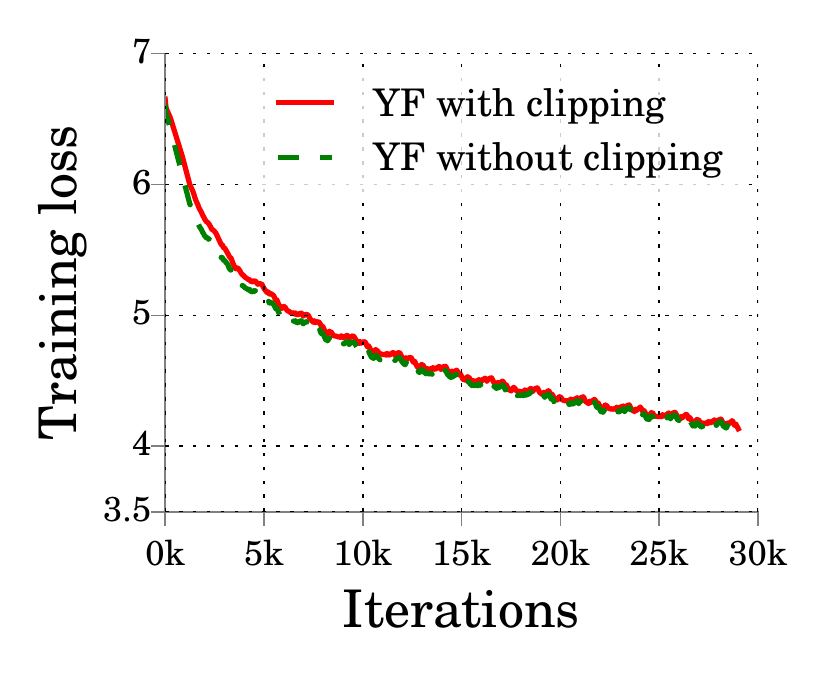} &
	\includegraphics[width=0.35\linewidth]{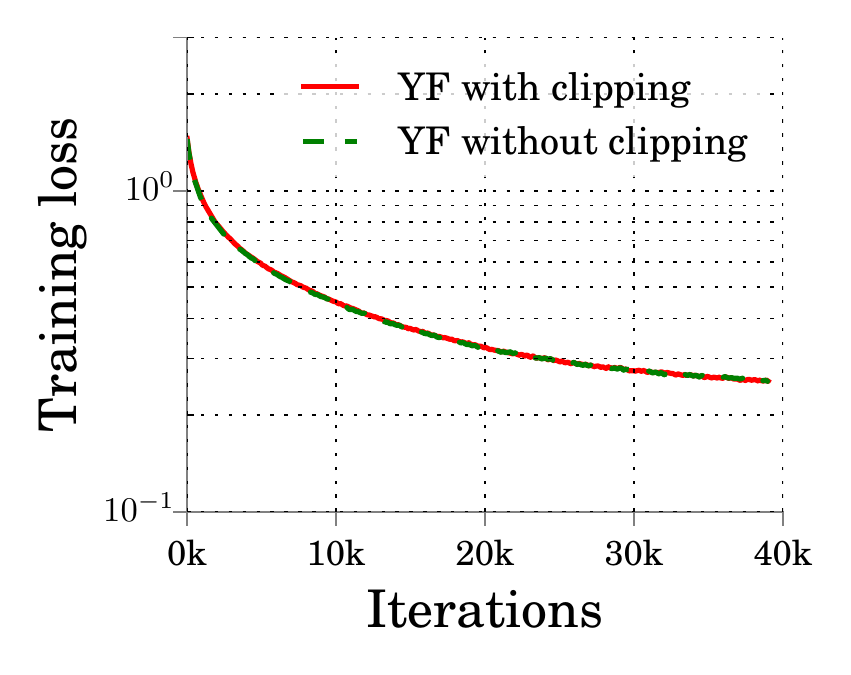}
\end{tabular}
\caption{Training losses on PTB LSTM (left) and CIFAR10 ResNet (right) for YellowFin with and without adaptive clipping.}
\label{fig:infl_clip}
\end{figure}

In Section~\ref{sec:stability}, we saw that adaptive clipping stabilizes the training on objectives that exhibit exploding gradients. In Figure~\ref{fig:infl_clip}, we demonstrate that the adaptive clipping does not hurt performance on models that do not exhibit instabilities without clipping. Specifically, for both PTB LSTM and CIFAR10 ResNet, the difference between \tuner with and without adaptive clipping diminishes quickly.  
\label{sec:infl_clip}

%% file: async_app.tex
\section{Closed-loop \tuner for asynchronous training}
\label{sec:async_app}
In Section~\ref{sec:async_tuner}, we briefly discuss the closed-loop momentum control mechanism in \asynctuner. In this section, after presenting more preliminaries on asynchrony, we show with details on the mechanism: 
it measures the dynamics on a running system and controls momentum with a negative feedback loop.
\paragraph{Preliminaries}
Asynchrony is a popular parallelization technique \citep{recht2011hogwild} that avoids synchronization barriers.
When training on $M$ asynchronous workers, staleness (the number of model updates between a worker's read and write operations) is on average $\tau=M-1$,
i.e., the gradient in the SGD update is delayed by $\tau$ iterations as $\nabla f_{S_{t - \tau}}(x_{t - \tau} )$.
Asynchrony yields faster steps, but can
increase the number of iterations to achieve the same solution,
a tradeoff between hardware and statistical 
efficiency~\citep{DBLP:journals/pvldb/ZhangR14}.
\citet{mitliagkas2016asynchrony} interpret asynchrony as added momentum dynamics.
Experiments in \citet{hadjis2016omnivore} support this finding, and demonstrate that reducing algorithmic momentum can compensate for asynchrony-induced momentum
and significantly reduce the number of iterations for convergence.
Motivated by that result, we use the model
in~\eqref{equ:exp_async_update_app}, where the total momentum, $\mu_T$, includes both asynchrony-induced and algorithmic  momentum, $\mu$, in~\eqref{eqn:momentum_gd}.
\begin{equation}
	\mathbb{E}[ x_{t+1} - x_t ] 
	= \mu_T \mathbb{E}[x_t - x_{t-1}] - \alpha \mathbb{E}\nabla f(x_{t})
\label{equ:exp_async_update_app}
\end{equation}
We will use this expression to design an estimator for the value of total momentum, $\hat{\mu}_T$.
This estimator is a basic building block of \asynctuner, that {\em removes the need to manually compensate for the effects of asynchrony}.

\paragraph{Measuring the momentum dynamics}
\Asynctuner estimates total momentum $\mu_{T}$ on a running system and uses a negative feedback loop to adjust algorithmic momentum accordingly.
Equation~\eqref{equ:exp_async_update} gives an estimate of $\hat{\mu}_T$ on a system with staleness $\tau$, based on \eqref{equ:exp_async_update}.
\begin{align}
\hat{\mu}_T
					= \mathop{\mathsf{median}}\left(
							\frac{x_{t - \tau} - x_{t - \tau-1} + \alpha \nabla_{S_{t-\tau -1}} f(x_{t - \tau - 1} )}
							{x_{t - \tau-1} - x_{t - \tau-2}}
					\right)
\label{eqn:momentum_measurement}
\end{align}
We use $\tau$-stale model values to match the staleness of the gradient,  and perform all operations in an elementwise fashion. 
This way we get a total momentum measurement from each variable; 
the median combines them into a more robust estimate.

\paragraph{Closing the asynchrony loop}
Given a reliable measurement of $\mu_{T}$, 
we can use it to adjust the value of algorithmic momentum so that the total momentum matches the \emph{target momentum} as decided by \tuner in Algorithm~\ref{alg:basic-algo}.
\Asynctuner in Algorithm~\ref{alg:async-algo} 
uses a simple negative feedback loop to achieve the adjustment.

\begin{algorithm}[h]
	\caption{\Asynctuner}
	\begin{algorithmic}[1]
	\State Input: $\mu\gets0$, $\alpha \gets 0.0001$, $\gamma\gets0.01, \tau$ (staleness)
	\For { $t\gets1$ to $T$}
	\State $x_t\!\gets\!x_{t - 1} + \mu (x_{t - 1} - x_{t - 2} ) - \alpha \nabla_{S_t} f(x_{t - \tau - 1} )$
	\State $\mu^*,\alpha \gets \Call{\tuner}{\nabla_{S_t} f(x_{t - \tau - 1} ), \beta}$ 
	\State $\hat{\mu_T} 
					\gets \mathop{\mathsf{median}}\left(
							\frac{x_{t - \tau} - x_{t - \tau-1} + \alpha \nabla_{S_{t-\tau-1}} f(x_{t - \tau - 1} )}
							{x_{t - \tau-1} - x_{t - \tau-2}}
					\right)$ \Comment{Measuring total momentum}
	\State $\mu \leftarrow \mu + \gamma \cdot (\mu^* - \hat{\mu_T})$ \Comment{Closing the loop}
	\EndFor
\end{algorithmic}
\label{alg:async-algo}
\end{algorithm}

%% file: model_spec.tex
\section{Model specification}
\label{sec:model_spec}
The model specification is shown in Table~\ref{tab:model_specification} for all the experiments in Section~\ref{sec:experiments}. 
CIRAR10 ResNet uses the regular ResNet units while CIFAR100 ResNet uses the bottleneck units. Only the convolutional layers are shown with filter size, filter number as well as the repeating count of the units. The layer counting for ResNets also includes batch normalization and Relu layers. The LSTM models are also diversified for different tasks with different vocabulary sizes, word embedding dimensions and number of layers.
\begin{table}
\vspace{1em}
\begin{scriptsize}
\centering
	\begin{tabular}{c@{\hskip 0.1in} c@{\hskip 0.1in} c@{\hskip 0.1in} c@{\hskip 0.1in} c@{\hskip 0.1in} c}
	\toprule
	network & \# layers & Conv 0 & Unit 1s & Unit 2s & Unit 3s \\
	\midrule
	CIFAR10 ResNet & 110 
	& $\left[\begin{array}{c c} 3 \times 3, & 4 \end{array} \right] $
	& $\left[\begin{array}{c c} 3 \times 3, & 4  \\ 3\times 3, & 4\end{array} \right]\times 6 $ 
	& $\left[\begin{array}{c c} 3 \times 3, & 8  \\ 3\times 3, & 8\end{array} \right]\times 6 $
	& $\left[\begin{array}{c c} 3 \times 3, & 16  \\ 3\times 3, & 16\end{array} \right]\times 6 $
	\\
	\midrule
	CIFAR100 ResNet & 164 
	& $\left[\begin{array}{c c} 3 \times 3, & 4 \end{array} \right] $
	& $\left[\begin{array}{c c} 1 \times 1, & 16  \\ 3\times 3, & 16 \\ 1 \times 1, & 64  \end{array} \right]\times 6  $
	& $\left[\begin{array}{c c} 1 \times 1, & 32  \\ 3\times 3, & 32 \\ 1 \times 1, & 128  \end{array} \right]\times 6  $
	& $\left[\begin{array}{c c} 1 \times 1, & 64  \\ 3\times 3, & 64 \\ 1 \times 1, & 256  \end{array} \right]\times 6  $ 
		\\
	\midrule
	\midrule
	network & \# layers & Word Embed. & Layer 1 & Layer 2 & Layer 3 \\
	\midrule
	TS LSTM & 2 & [65 vocab, 128 dim] & 128 hidden units & 128 hidden units & --  \\
	\midrule
	PTB LSTM & 2 & [10000 vocab, 200 dim] & 200 hidden units & 200 hidden units & -- \\
	\midrule
	WSJ LSTM & 3 & [6922 vocab, 500 dim] & 500 hidden units & 500 hidden units & 500 hidden units\\
	\bottomrule
	\end{tabular}
\end{scriptsize}
\caption{Specification of ResNet and LSTM model architectures.}
\label{tab:model_specification}
\end{table}

%% file: exp_spec.tex
\section{Specification for synchronous experiments}
\label{sec:exp_spec}
In Section~\ref{subsec:sync_exp}, we demonstrate the synchronous experiments with extensive discussions. 
For the reproducibility, we provide here the specification of learning rate grids. The number of iterations as well as epochs, i.e. the number of passes over the full training sets, are also listed for completeness. For \tuner in all the experiments in Section~\ref{sec:experiments}, we uniformly use sliding window size $20$ for extremal curvature estimation and $\beta = 0.999$ for smoothing. For momentum SGD and Adam, we use the following configurations.
\begin{itemize}
	\item CIFAR10 ResNet
		\begin{itemize}
			\item $40$k iterations (${\sim} 114$ epochs)
			\item Momentum SGD learning rates $\{0.001, 0.01 \text{(best)}, 0.1, 1.0\}$, momentum 0.9
			\item Adam learning rates $\{0.0001, 0.001 \text{(best)}, 0.01, 0.1\}$
		\end{itemize}
	\item CIFAR100 ResNet
		\begin{itemize}
			\item $120$k iterations (${\sim} 341$ epochs)
			\item Momentum SGD learning rates $\{0.001, 0.01 \text{(best)}, 0.1, 1.0\}$, momentum 0.9
			\item Adam learning rates $\{0.00001, 0.0001\text{(best)}, 0.001, 0.01\}$
		\end{itemize}
	\item PTB LSTM
		\begin{itemize}
			\item 30k iterations (${\sim} 13$ epochs)
			\item Momentum SGD learning rates $\{0.01, 0.1, 1.0 \text{(best)}, 10.0\}$, momentum 0.9
			\item Adam learning rates $\{0.0001, 0.001 \text{(best)}, 0.01, 0.1\}$
		\end{itemize}
	\item TS LSTM
		\begin{itemize}
			\item ${\sim}21$k iterations ($50$ epochs)
			\item Momentum SGD learning rates $\{0.05, 0.1, 0.5, 1.0 \text{(best)}, 5.0\}$, momentum 0.9
			\item Adam learning rates $\{0.0005, 0.001, 0.005 \text{(best)}, 0.01, 0.05\}$
			\item Decrease learning rate by factor 0.97 every epoch for all optimizers, following the design by~\citet{karpathy2015visualizing}.
		\end{itemize}
	\item WSJ LSTM
		\begin{itemize}
			\item ${\sim} 120$k iterations ($50$ epochs)
			\item Momentum SGD learning rates $\{0.05, 0.1, 0.5 \text{(best)}, 1.0, 5.0\}$, momentum 0.9
			\item Adam learning rates $\{0.0001, 0.0005, 0.001 \text{(best)}, 0.005, 0.01\}$
			\item Vanilla SGD learning rates $\{0.05, 0.1, 0.5, 1.0 \text{(best)}, 5.0\}$
			\item Adagrad learning rates $\{0.05, 0.1, 0.5 (\text{best}), 1.0, 5.0\}$
			\item Decrease learning rate by factor 0.9 every epochs after 14 epochs for all optimizers, following the design by~\citet{charniakparsing}.
		\end{itemize}
\end{itemize}

\section{Additional experiment results}
\label{sec:add_exp}
\subsection{Training losses on CIFAR10 and CIFAR100 ResNet}
\begin{figure}
\centering
	\begin{tabular}{c c}
		\includegraphics[width=0.4\linewidth]{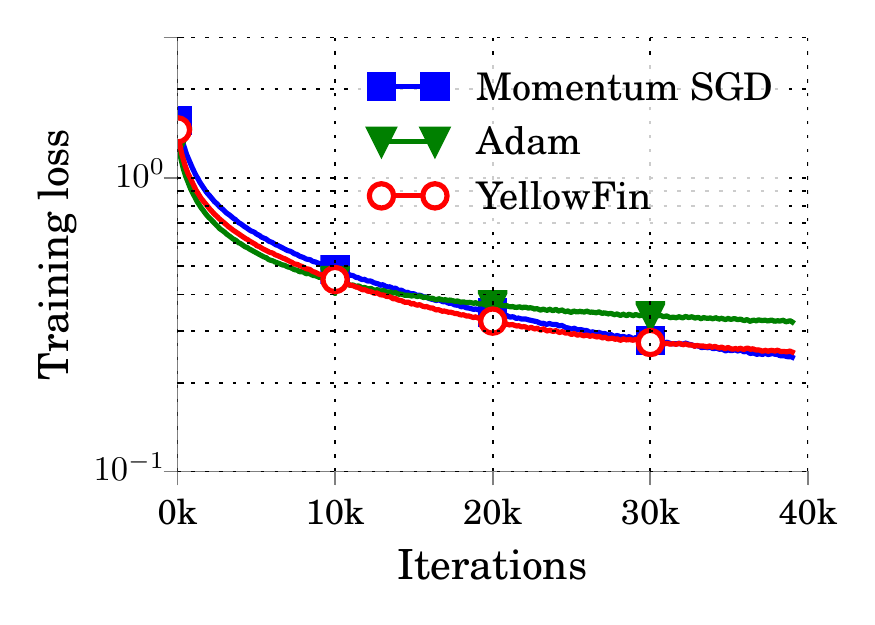} &
		\includegraphics[width=0.4\linewidth]{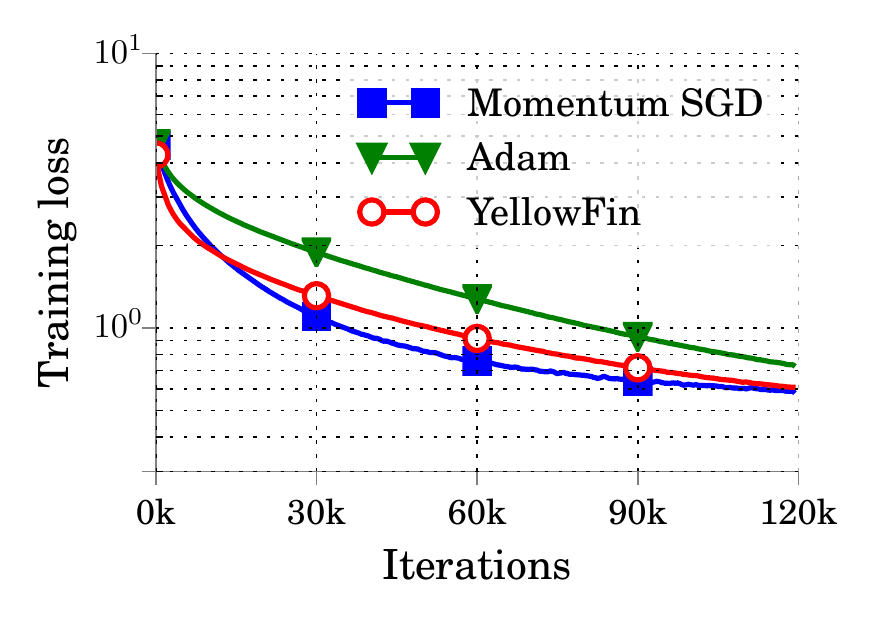}
	\end{tabular}
	\caption{
	Training loss for ResNet on 100-layer CIFAR10 ResNet (left) and 164-layer CIFAR100 bottleneck ResNet. }
	\label{fig:loss_result_cifar}
\end{figure}

In Figure~\ref{fig:loss_result_cifar}, we demonstrate the training loss on CIFAR10 ResNet and CIFAR100 ResNet. Specifically, \tuner can match the performance of hand-tuned momentum SGD, and achieves 1.93x and 1.38x speedup comparing to hand-tuned Adam respectively on CIFAR10 and CIFAR100 ResNet.

\subsection{Importance of momentum adaptivity}
\label{sec:importance_momentum}
\begin{figure}
\centering	
\begin{tabular}{c c}
	\includegraphics[width=0.4\linewidth]{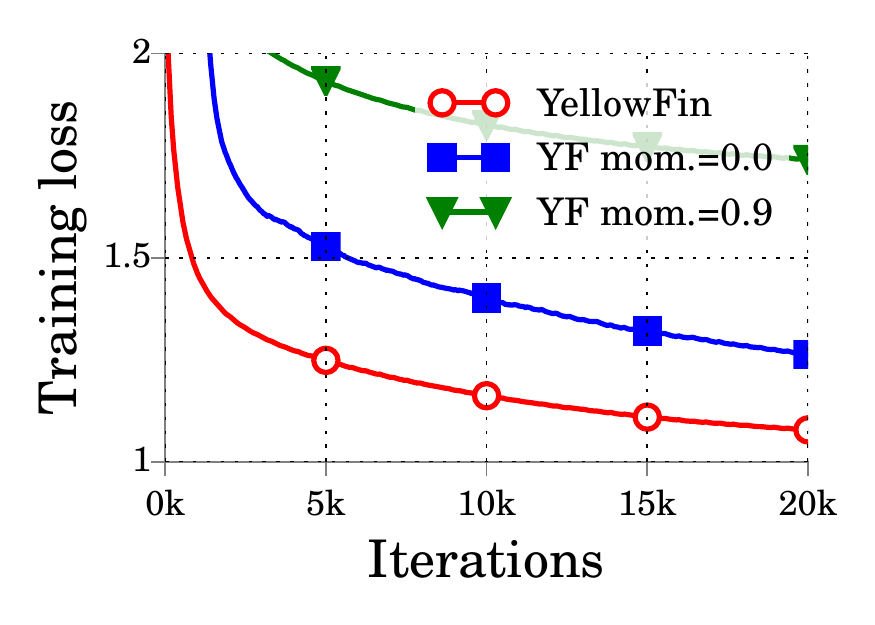} &
	\includegraphics[width=0.4\linewidth]{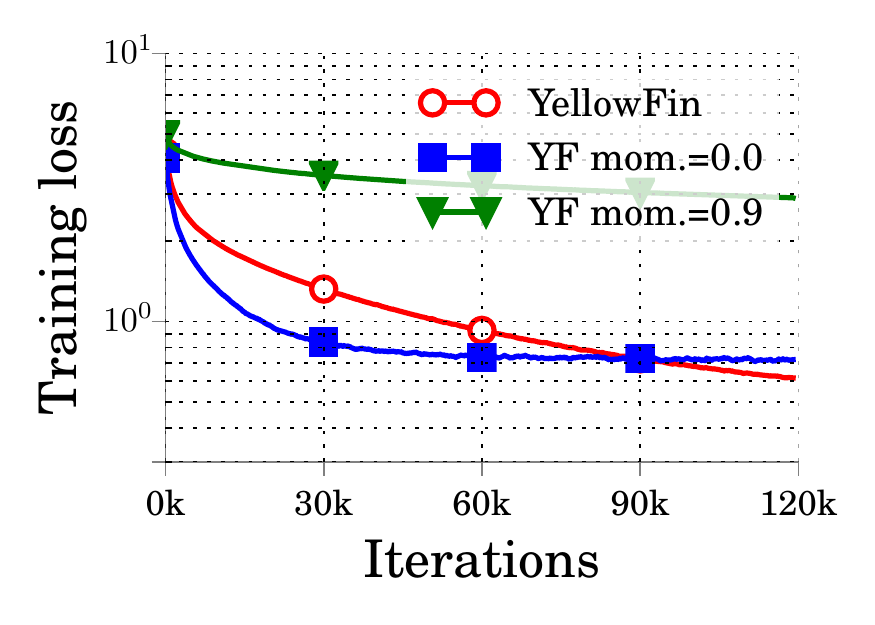}
\end{tabular}
\caption{Training loss comparison between \tuner with adaptive momentum and \tuner with fixed momentum value. This comparison is conducted on TS LSTM (left) and CIFAR100 ResNet (right).}
\label{fig:cmp_fix_mom}
\end{figure}

To further emphasize the importance of momentum adaptivity in \tuner, we run YF on CIFAR100 ResNet and TS LSTM. In the experiments, \tuner tunes the learning rate. Instead of also using the momentum tuned by YF, we continuously feed prescribed momentum value $0.0$ and $0.9$ to the underlying momentum SGD optimizer which YF is tuning. In Figure~\ref{fig:cmp_fix_mom}, when comparing to \tuner with prescribed momentum 0.0 or 0.9, \tuner with adaptively tuned momentum achieves observably faster convergence on both TS LSTM and CIFAR100 ResNet. It empirically demonstrates the essential role of momentum adaptivity in \tuner.

\subsection{Tuning momentum can improve Adam in async.-parallel setting}
\begin{wrapfigure}[13]{R}{0.425\linewidth}
\vspace{-3em}
\begin{minipage}{1.0\linewidth}
	\begin{figure}[H]
		\includegraphics[width=\linewidth]{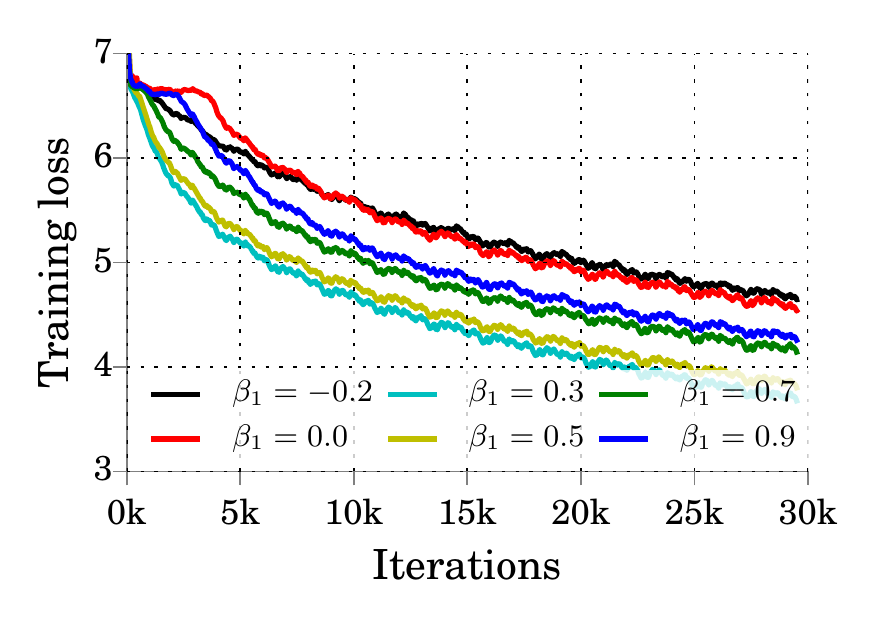}
			\vspace{-2em}
		\caption{Hand-tuning Adam's momentum under asynchrony.}
		\label{fig:adam_async_mom}
	\end{figure}
\end{minipage}	
\end{wrapfigure}
We conduct experiments on PTB LSTM with 16 asynchronous workers using Adam using the same protocol as in Section~\ref{sec:async_exp}.
Fixing the learning rate to the value achieving the lowest smoothed loss in Section~\ref{subsec:sync_exp}, we sweep the smoothing parameter $\beta_1$~\citep{kingma2014adam} of the first order moment estimate in grid $\{-0.2, 0.0, 0.3, 0.5, 0.7, 0.9\}$. $\beta_1$ serves the same role as momentum in SGD and we call it the momentum in Adam. Figure~\ref{fig:adam_async_mom} shows tuning momentum for Adam under asynchrony gives measurably better training loss. 
This result emphasizes the importance of momentum tuning in asynchronous settings and suggests that state-of-the-art adaptive methods can perform sub-optimally when using prescribed momentum.

\subsection{Accelerating \tuner with finer grain learning rate tuning}
\label{sec:boost_exp}
 As an adaptive tuner, \tuner does not involve manual tuning. It can present faster development iterations on model architectures than grid search on optimizer hyperparameters. In deep learning practice for computer vision and natural language processing, after fixing the model architecture, extensive optimizer tuning (e.g. grid search or random search) can further improve the performance of a model. A natural question to ask is can we also slightly tune \tuner to accelerate convergence and improve the model performance. Specifically, we can manually multiply a positive number, the learning rate factor, to the auto-tuned learning rate in \tuner to further accelerate. 
 
In this section, we empirically demonstrate the effectiveness of learning rate factor on a 29-layer ResNext (2x64d)~\citep{xie2016aggregated} on CIFAR10 and a Tied LSTM model~\citep{press2016using} with 650 dimensions for word embedding and two hidden units layers on the PTB dataset. 
 	 When running \tuner, we search for the optimal learning rate factor in grid $\{\frac{1}{3}, 0.5, 1, 2(\text{best for ResNext} ), 3 (\text{best for Tied LSTM} ), 10\}$. 
 	 Similarly, we search the same learning rate factor grid for Adam, multiplying the factor to its default learning rate $0.001$. 
 	 To further strengthen the performance of Adam as a baseline, we also run it on conventional logarithmic learning rate grid $\{5e^{-5}, 1e^{-4}, 5e^{-4}, 1e^{-3}, 5e^{-3}\}$ for ResNext and $\{1e^{-4}, 5e^{-4}, 1e^{-3}, 5e^{-3}, 1e^{-2}\}$ for Tied LSTM. We report the best metric from searching the union of learning rate factor grid and logarithmic learning rate grid as searched Adam results.
 	 Empirically, learning factor $\frac{1}{3}$ and $1.0$ works best for Adam respectively on ResNext and Tied LSTM. 
 	 
\begin{figure}
\centering
\begin{tabular}{c c} 
 	\includegraphics[width=0.4\linewidth]{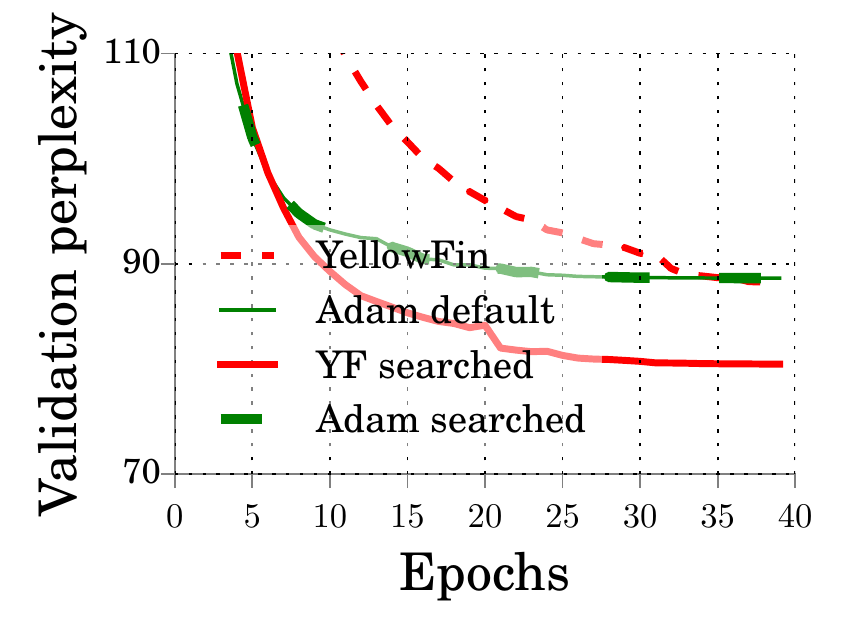} &
 	\includegraphics[width=0.4\linewidth]{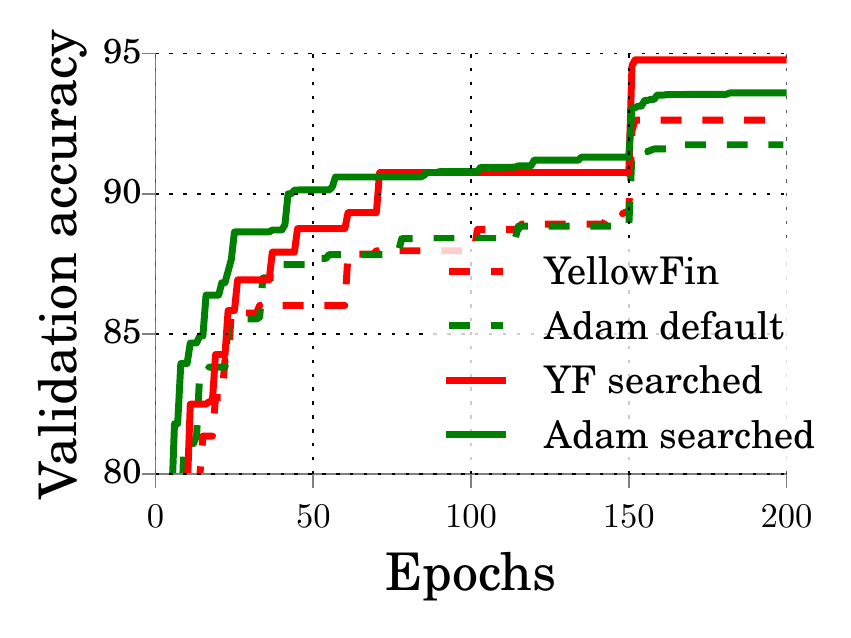}
\end{tabular}
\caption{Validation perplexity on Tied LSTM and validation accuracy on ResNext. Learning rate fine-tuning using grid-searched factor can further improve the performance of \tuner in Algorithm~\ref{alg:basic-algo}. \tuner with learning factor search can outperform hand-tuned Adam  on validation metrics on both models.}
\label{fig:yf_boost}
\end{figure}

As shown in Figure~\ref{fig:yf_boost}, with the searched best learning rate factor, \tuner can improve validation perplexity on Tied LSTM from $88.7$ to $80.5$, an improvement of more than $9\%$. Similarly, the searched learning rate factor can improve test accuracy from $92.63$ to $94.75$ on ResNext. More importantly, we can observe, with learning rate factor search on the two models, \tuner can achieve better validation metric than the searched Adam results. It demonstrates that finer-grain learning rate tuning, i.e. the learning rate factor search, can be effectively applied on \tuner to improve the performance of deep learning models.

